%% file: main.tex
\documentclass[sigconf]{acmart}
\usepackage{tcolorbox}

\usepackage{amsmath}
\usepackage{subcaption}
\usepackage{xcolor}
\usepackage{makecell}

\usepackage{fancyvrb}
\usepackage{framed}

\usepackage{verbatim}
\usepackage{listings}
\usepackage{algorithm}
\usepackage{algpseudocode}
\usepackage{enumitem} 
\usepackage{graphicx}
\usepackage{multirow}
\usepackage{array}
\usepackage{xcolor,colortbl}
\usepackage{longtable}

\usepackage{array}
\usepackage{caption}
\usepackage{makecell}

\usepackage{listings}
\usepackage{xcolor}

\makeatletter

\usepackage{fancyvrb}

\lstdefinestyle{llmstyle}{
    basicstyle=\ttfamily\footnotesize,
    breaklines=true,
    breakatwhitespace=false,
    columns=fullflexible,   
    keepspaces=true,
    showstringspaces=false,
    frame=none,
    xleftmargin=0pt,
    xrightmargin=0pt,
    breakindent=0pt,
    linewidth=\columnwidth,
    backgroundcolor=\color{white},   
    postbreak={},          
    language={},            
    mathescape=false,       
    tabsize=4,
    numbers=none,
    escapeinside={}         
}

\lstnewenvironment{llmprompt}
    {\lstset{style=llmstyle}}
    {}

\definecolor{codegreen}{rgb}{0,0.6,0}
\definecolor{codegray}{rgb}{0.5,0.5,0.5}
\definecolor{codepurple}{rgb}{0.58,0,0.82}
\definecolor{backcolour}{rgb}{0.95,0.95,0.92}

\lstdefinestyle{mystyle}{
    backgroundcolor=\color{backcolour},   
    commentstyle=\color{codegreen},
    keywordstyle=\color{magenta},
    numberstyle=\tiny\color{codegray},
    stringstyle=\color{codepurple},
    basicstyle=\ttfamily\small,  
    breakatwhitespace=false,         
    breaklines=true,                 
    captionpos=b,                    
    keepspaces=true,                 
    numbers=left,                    
    numbersep=5pt,                  
    showspaces=false,                
    showstringspaces=false,
    showtabs=false,                  
    tabsize=2
}
\AtBeginDocument{%
  }

\copyrightyear{2025}
\acmYear{2025}
\setcopyright{cc}
\setcctype{by}
\acmConference[CHI '25]{CHI Conference on Human Factors in Computing Systems}{April 26-May 1, 2025}{Yokohama, Japan}
\acmBooktitle{CHI Conference on Human Factors in Computing Systems (CHI '25), April 26-May 1, 2025, Yokohama, Japan}\acmDOI{10.1145/3706598.3713675}
\acmISBN{979-8-4007-1394-1/25/04}

\usepackage{acmart-taps}

\begin{document}

\title{Plurals: A System for Guiding LLMs Via Simulated Social Ensembles}

\author{Joshua Ashkinaze}
\affiliation{%
  \institution{University of Michigan}
  \country{United States}}
\email{jashkina@umich.edu}

\author{Emily Fry}
\affiliation{%
 \institution{Oakland Community College}
  \institution{University of Michigan}
  \country{United States}}
\email{exfry@student.oaklandcc.edu}

\author{Narendra Edara}
\affiliation{%
  \institution{University of Michigan}
  \country{United States}}
\email{nedara@umich.edu}

\author{Eric Gilbert}
\affiliation{%
  \institution{University of Michigan}
  \country{United States}}
\email{eegg@umich.edu}

\author{Ceren Budak}
\affiliation{%
  \institution{University of Michigan}
  \country{United States}}
\email{cbudak@umich.edu}

\renewcommand{\shortauthors}{Ashkinaze et al.}

\begin{abstract}
\looseness-1 Recent debates raised concerns that language models may favor certain viewpoints. But what if the solution is not to aim for a ``view from nowhere'' but rather to leverage different viewpoints? We introduce Plurals, a system and Python library for pluralistic AI deliberation. Plurals consists of Agents (LLMs, optionally with personas) which deliberate within customizable Structures, with Moderators overseeing deliberation. Plurals is a generator of simulated social ensembles. Plurals integrates with government datasets to create nationally representative personas, includes deliberation templates inspired by deliberative democracy, and allows users to customize both information-sharing structures and deliberation behavior within Structures. Six case studies demonstrate fidelity to theoretical constructs and efficacy. Three randomized experiments show simulated focus groups produced output resonant with an online sample of the relevant audiences (chosen over zero-shot generation in 75\% of trials). Plurals is both a paradigm and a concrete system for pluralistic AI. 

\end{abstract}



\maketitle

\section{Introduction}

\enlargethispage{10pt}
\begin{figure*}
 \includegraphics[width=\textwidth]{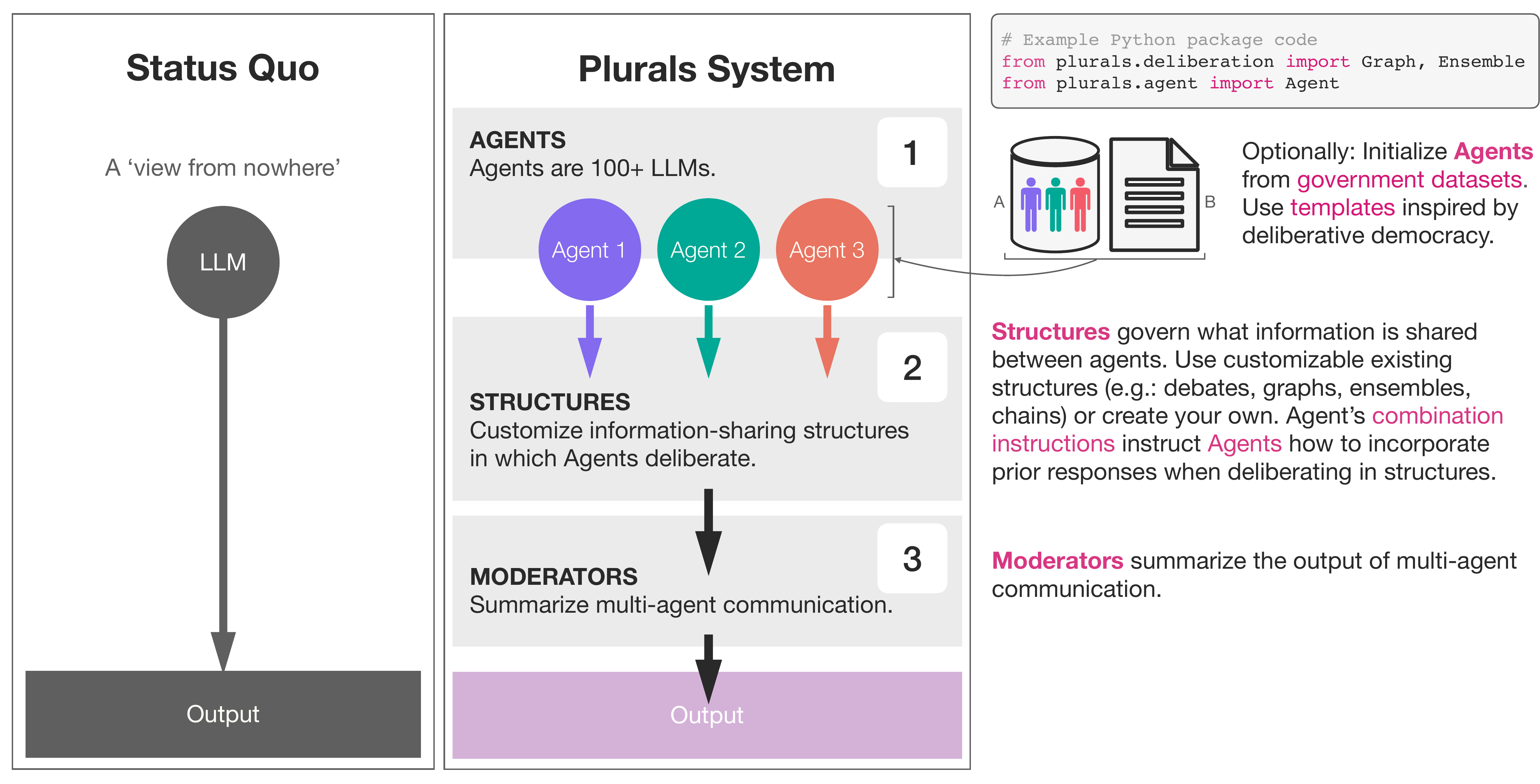}
  \caption{System diagram of Plurals---an end-to-end generator of simulated social ensembles. (1) \textbf{Agents} complete tasks within (2) \textbf{Structures}, with communication optionally summarized by (3) \textbf{Moderators}. Plurals integrates with government datasets (1a) and templates inspired by deliberative democracy theory (1b). The building block is Agents, which are large language models (LLMs) that have system instructions and tasks. System instructions can be generated from user input, government datasets (American National Election Studies; ANES), and templates from deliberative democracy literature~\cite{bachtiger_deliberative_2018}. Agents exist within Structures, which define what information is shared. Combination instructions tell Agents how to combine the responses of other Agents when deliberating in the Structure. Users can customize an Agent's combination instructions or use existing templates drawn from deliberation literature and beyond. Moderators aggregate responses from multi-agent deliberation.}
  \Description{System diagram of Plurals. The system consists of three components: Agents are language models that perform tasks; Structures define information sharing between agents; Moderators summarize agent communication. The visual diagram depicts each of these components and also contrasts the multi-perspective approach of Plurals with the conventional approach of a ``view from nowhere''.}
  \label{fig:teaser}
\vspace{-10pt}
\end{figure*}

There is a fundamental tension between how generative AI models are built and how they are used. Companies typically build a small number of foundation or ``generalist'' models that dominate the market~\cite{vipra_market_2023}. However, these generalist models are used by a diverse base of users---with varying preferences and values. Invariably, this tension sparked allegations of bias, with supposedly neutral models accused of favoring certain viewpoints~\cite{feng_pretraining_2023, braun_politics_2023, fujimoto_revisiting_2023}. 

While a tempting solution is to aim for models that have ``no bias'' and hold a ``view from nowhere''~\cite{haraway_situated_1997}, truly neutral models are likely infeasible. Some scholars argue that all knowledge is situated~\cite{haraway_situated_1997}. But with open-ended text generation, defining some unbiased ground truth is especially difficult. For many use cases, there is no unbiased ground truth. This difficulty is compounded by the fact that users can ask models a large variety of questions. Any bias benchmark can only capture an infinitesimal slice of the query space~\cite{raji_ai_2021}.

As a motivating example, imagine a company preparing to launch a new work-from-home policy. The CEO seeks to determine which aspects of the policy memo will raise concerns for employees. Or suppose a housing justice group aims to identify the most effective messaging for a homeless shelter proposal. LLMs can theoretically be deployed for both cases. But what viewpoint should the LLM adopt? Different employees and residents have different perspectives. The standard approach of prompting a single model is unlikely to represent diverse viewpoints. We propose an alternative approach: A system of LLMs engage in controlled deliberation, simulating distinct viewpoints. The CEO could create a network of simulated employees to provide feedback, upweighting the voices of the most affected groups. The housing justice group could create a sequence of LLMs with demographically weighted personas to provide iterative feedback based on preceding concerns.

As an alternative to ``bias-free'' models, we introduce a new pluralistic AI system~\cite{sorensen_roadmap_2024}, Plurals, that can accomplish these tasks. It is a public-facing Python library (\autoref{fig:teaser} for system overview, \autoref{fig:2x3panel} for code snippets, see here\footnote{\href{https://github.com/josh-ashkinaze/plurals}{https://github.com/josh-ashkinaze/plurals}}  \enlargethispage{10pt}
 for library). Plurals consists of Agents (optionally integrated with government datasets for nationally representative personas) which deliberate within customizable Structures, with Moderators overseeing deliberation. Plurals is an end-to-end generator of customizable ``simulated social ensembles''. We incorporate interaction templates inspired by deliberative democracy theory and integration with government datasets for nationally representative personas. For example, to create an Agent representing a male California resident, our system samples a statistically representative citizen from American National Election Studies, and then uses the citizen's demographics and political stances as an LLM prompt. We draw on deliberative democracy theory, which emphasizes dialogue between different views~\cite{bachtiger_deliberative_2018, marti_pluralism_2017}, as a blueprint. Our work builds on research in deliberation~\cite{bachtiger_deliberative_2018, morrell_listening_2018, brown_deliberation_2018, smith_mini-publics_2018, fraser_rethinking_1990, habermas_structural_1991, marti_pluralism_2017}, pluralistic sociotechnical systems~\cite{argyle_out_2023, lee_webuildai_2019, gordon_jury_2022, zhang_policykit_2020}, and multi-agent AI alignment approaches~\cite{vahidov_pluralistic_2004,irving_ai_2018, talebirad_multi-agent_2023, kim_can_2024, chan_chateval_2023}. To our knowledge, Plurals is the first general-purpose toolkit for pluralistic, multi-agent interactions modeled after deliberative democracy.

We conducted six empirical case studies of Plurals' theoretical fidelity and efficacy. Across three randomized experiments, we find that Plurals can simulate focus groups, leading to output that resonates with an online sample of the relevant audiences (above zero-shot and chain-of-thought generation). We view Plurals as a toolkit for building towards pluralistic artificial intelligence. This work has three contributions:

\begin{itemize}
    \item \textbf{Theoretical}: We created a multi-agent system incorporating ideals of deliberative democracy theory. Our system also introduces ``interactional pluralism'', a pluralism that exists not only in the distribution of agent properties but also in the protocols governing their interactions. Users can customize how Agents should combine information with each other and the information-sharing structures in which Agents exist. 

    \item \textbf{System}: Plurals is a public-facing Python package with documentation and tutorials. We made these theoretical ideals concrete, creating a usable system for pluralistic AI. 
    
    \item \textbf{Empirical}: We present early empirical results from our system. Two case studies demonstrate \textit{mechanistic fidelity}, that the system is doing what we claim it is doing. Three case studies demonstrate \textit{efficacy}: Simulated focus groups of liberals and conservatives yield output that is compelling to real liberals and conservatives. One case study also shows how Plurals can be used as a programmable environment for creating guardrails. 
\end{itemize}

We provide an overview of the system (\autoref{overview}), review its grounding in prior work (\autoref{grounding}), explain its principles (\autoref{principles}), and describe it in detail (\autoref{details}) with code snippets. We then present six empirical case studies demonstrating theoretical fidelity and efficacy (\autoref{case_studies}). We discuss limitations (e.g.: fidelity, steerability; \autoref{limitations}) and ethical considerations (\autoref{ethics}). We conclude with future research directions and broader implications (\autoref{discussion}).

\subsection{Brief System Overview}
\label{overview}
Plurals allows users to create simulated social ensembles with \textbf{Agents}, \textbf{Structures}, and \textbf{Moderators}: Agents complete tasks within Structures, which define how information is shared between Agents. Moderators can summarize multi-agent communication. Each abstraction is highly customizable. Agents can use various LLMs and have system instructions set manually, through persona generation, or via American National Election Studies (ANES) integration. Structures vary in information-sharing, complexity, and randomness. For example, users can define custom networks of Agents in a few lines of code (\autoref{fig:2x3panel}). The behavior of Agents within Structures (how they should combine information from other agents) can be tuned via combination instructions. Our package comes pre-populated with templates for personas, combination instructions, and moderators---drawing on deliberative democracy theory and prior work. 

\section{System Grounding}

\label{grounding}

Plurals is grounded in deliberation literature, sociotechnical systems that broaden technological perspectives, and multi-agent systems for AI alignment. The result is an end-to-end generator of simulated social ensembles---groups that engage in deliberation. We integrate deliberative theory into our system by incorporating templates of first- and second-generation deliberative ideals and using deliberative theory to inform the structure of AI deliberation. We extend previous work on broadening technological perspectives, such as Argyle et al.'s dataset-based personas~\cite{argyle_out_2023}, Gordon et al.'s ``juries''~\cite{gordon_jury_2022}, and Zhang et al.'s PolicyKit~\cite{zhang_policykit_2020}. Our system encompasses individual, group, and governance-level simulations, unlike previous approaches that focused on flexibility at only one of these three levels. By drawing on the concept of deliberative ``mini-publics'' (groups who engage in deliberation~\cite{smith_mini-publics_2018}), we evolve from aggregative methods (like juries) to a more deliberative approach. Additionally, we contribute to multi-agent AI research by offering a flexible system for creating diverse interaction structures and providing a reusable infrastructure for experiments.

\subsection{Deliberation}
\label{deliberation_rr}

Deliberation is defined as ``mutual communication that involves weighing and reflecting on preferences, values, and interests regarding matters of common concern''~\cite{bachtiger_deliberative_2018}. As Bächtiger et al. distinguish~\cite{bachtiger_deliberative_2018}, deliberative democracy differs from aggregative democracy. The former centers talking and the latter centers voting---though they can co-occur (e.g., talking before voting~\cite{fishkin_consulting_2003, isernia_europolis_2014}). Deliberation occurs in many different forms, in many different ways, and has many different outcome measures. In what follows, we clarify the aspects of deliberation literature that inform our system. 

\paragraph{Practice of Deliberation} The abstractions of Plurals map to the \textit{practice} of deliberation. Ryfe breaks deliberative practice into three phases~\cite{ryfe_does_2005}: (1) The organization of the encounter, (2) the deliberation within the encounter, and (3) the final product of deliberation. Agents are the building blocks of deliberation. As such, Agent initialization corresponds to Phase 1. The deliberation within the encounter is governed by Structures and combination instructions, corresponding to Phase 2. Finally, Moderators can amend the final product of deliberation, corresponding to Phase 3. Separate from Ryfe, Morrell~\cite{morrell_listening_2018} considers three factors of deliberation that affect outcomes: individual dispositions, institutional structures, and facilitators/moderators. Again, these correspond almost directly to our abstractions of Agents, Structures, and Moderators. More generally, formal deliberation nowadays often occurs in ``mini-publics''~\cite {smith_mini-publics_2018}. These are groups of citizens who come together to deliberate, often in an advisory role. Plurals is an end-to-end generator of simulated social ensembles. This is analogous to reproducing the process of mini-public deliberation. 

\paragraph{Deliberative Ideals} While the ideals of deliberation are not universally agreed upon, we adopt the dichotomy of ``first-generation'' and ``second-generation'' ideals articulated by Bächtiger et al.~\cite{bachtiger_deliberative_2018}. According to Bächtiger et al., the first generation of deliberative theorists (e.g., Habermas~\cite{habermas_structural_1991}) emphasized rationality, achieving a universal consensus, and reason-giving. The second generation of deliberative theorists took a more expansive view of deliberation, beyond rationality and universalism~\cite{bachtiger_deliberative_2018}. For example, second-wave deliberation also valued more emotional forms of communication~\cite{neblo_impassioned_2020}, lived experience, testimony, and storytelling~\cite{bachtiger_deliberative_2018}. Furthermore, to second-wave theorists, the goal was not \textit{necessarily} a universal consensus (since legitimate disagreement may still exist after perfect deliberation~\cite{marti_pluralism_2017}), but rather a clarifying of understanding~\cite{bachtiger_deliberative_2018, fraser_rethinking_1990}. 

We incorporate these ideals into our system as both persona templates (how LLMs should enact personas) and combination instructions (how LLMs should combine information with others). To do this, we started with the taxonomy of first-generation and second-generation principles from~\cite{bachtiger_deliberative_2018}. Two authors then engaged in an iterative, two-step process where we first decided whether each dimension was relevant to AI agents, and then how to operationalize this dimension for both generations of deliberation thought. Appendix \autoref{tab:deliberation_ideals} lists how we operationalized each ideal. 

Some, but not all, ideals or benefits of human deliberation may apply to AI deliberation. Deliberative mini-publics can be useful for the outcomes that they produce~\cite{warren_deliberative_1996, bachtiger_deliberative_2018, smith_mini-publics_2018} or the process that produces these outcomes. Regarding the latter, deliberation proponents argue deliberation has certain \textit{epistemic} (outcome-independent) benefits---such as increased perceived legitimacy for decisions when the sequence of thought leading to them is made public~\cite{estlund_epistemic_2018}. It is the former---outcome-oriented benefits---that is relevant to AI deliberation.

To be clear, our system is \textit{inspired} by human deliberation; it is not meant to substitute for it. By analogy, engineers often draw on the natural world to create artifacts. For example, Velcro was inspired by burrs sticking to the inventor's dog~\cite{hook_and_loop_invention_nodate}. The limits of human deliberation as a metaphor are discussed in~\autoref{imperfect_metaphor}.

\subsection{Pluralistic Sociotechnical Systems}

Other projects have sought to broaden the representation of technology, what we term ``pluralistic sociotechnical systems'' for shorthand. These approaches usually focus exclusively on individuals~\cite{argyle_out_2023}, groups~\cite{gordon_jury_2022, lee_webuildai_2019}, or governance structures~\cite{zhang_policykit_2020}. As an end-to-end generator of simulated social ensembles, Plurals does all three. 

Our system extends prior work aimed at broadening the representation of technological systems through simulation. These approaches address the inherent problems of collapsing diverse viewpoints into a single perspective, a phenomenon we term ``output collapse''. In data labeling, annotators often disagree~\cite{sandri_why_2023, popovic_agree_2021, miceli_between_2020}, yet traditional supervised learning typically resolves these disparities by selecting the majority label. This majority-driven approach can silence minority viewpoints or result in a system that behaves like a ``pseudo-human''~\cite{gordon_jury_2022}, presenting a blurred representation that diverges from individual perspectives.

To address output collapse, researchers developed systems that simulate specific perspectives~\cite{davani_dealing_2022, lee_webuildai_2019, he_cura_2023, gordon_jury_2022}. Plurals follows this tradition. The most similar system is Juries~\cite{gordon_jury_2022}, which is an architecture and interface for person-specific models. Juries allows end-users to create panels of simulated annotators who make classifications, with the option to upweight dissenting voices. 

While the above work primarily addresses individuals or small groups, some systems enhance technology's representativeness by customizing governance structures. For example, PolicyKit~\cite{zhang_policykit_2020} allows online communities to create arbitrary governance structures easily, essentially letting communities embed their own values. Similarly, Schneider et al. created ``modular politics''~\cite{schneider_modular_2021}, where communities construct governance structures from distinct components. 

Large language models (LLMs) have intensified both the problem of output collapse and the potential solutions to combat it. While a single ``ground truth'' was often contested in conventional classification~\cite{sandri_why_2023, popovic_agree_2021}, open-ended text generation further complicates the notion of a single, ``correct'' answer. Simultaneously, LLMs can \textit{potentially} be steered to adopt viewpoints through ``personas''~\cite{jiang_personallm_2024, salminen_deus_2024, ha_clochat_2024}. We adopt Argyle et al.'s~\cite{argyle_out_2023} method of generating personas from government datasets to use as LLM prompts. Both Argyle et al.'s method~\cite{argyle_out_2023} and Gordon et al.'s Juries~\cite{gordon_jury_2022} employ multiple individual characteristics to construct personas. By using nationally representative datasets, we create personas reflecting general population attributes. These intersectional personas should theoretically enhance diversity beyond single-attribute personas, reducing homogenization (Case Study 1).

Plurals is an evolution and extension of the above ideas. Plurals is an evolution of prior work: What Juries is to aggregative democracy, Plurals is to deliberative democracy~\cite{tsai_generative_2024}. Unlike Juries' focus on classification labels, Plurals can generate open-ended text. As Gordon et al.~\cite{gordon_jury_2022} write, ``jury learning does not draw on the deliberative nature of juries, which has been the subject of decades of study in legal literature.'' This deliberation is our contribution. Plurals also expands the core idea of Juries. In Plurals syntax (\autoref{fig:2x3panel}), a Gordon et al. jury is an \verb|ensemble| where Agents complete tasks in parallel without information sharing. This is just one communication structure. By allowing users to create diverse structures and customize Agent deliberation within these structures, we offer a more comprehensive approach to studying and implementing pluralistic AI. Finally, unlike Juries~\cite{gordon_jury_2022}, Plurals does not require a task-specific representative dataset with annotator demographics (which can be prohibitive to obtain). By allowing users to change governance structures, Plurals is conceptually similar to PolicyKit. However, Plurals differs from PolicyKit in that Plurals supports the construction of Agents and Moderators (the ``before'' and ``after'' of Structures using Ryfe's three-part terminology of deliberation~\cite{ryfe_does_2005}). In brief, Plurals allows end-to-end generation of simulated social ensembles.

\subsection{Multi-Agent Systems for AI Alignment}
Multi-agent systems have a long history in artificial intelligence~\cite{oliveira_multi-agent_1999, van_der_hoek_chapter_2008}. Now there is substantial interest in multi-agent LLM systems~\cite{pang_self-alignment_2024, mangal_coalitions_2024, hu_agentscomerge_2024, hua_assistive_2024, tsao_multi-agent_2023, ni_mechagents_2024, khan_debating_2024}. Our system incorporates aspects of these systems such as debate~\cite{irving_ai_2018} and the idea of role-based communication~\cite{pang_self-alignment_2024, zhu_role-based_2008}. 

Like our system, several multi-agent systems are explicitly designed with the goal of alignment~\cite{irving_ai_2018, pang_self-alignment_2024, mangal_coalitions_2024, hu_agentscomerge_2024}. Broadly, these systems typically center interactions between agents or agent roles. For example, several projects have explored the role of AI alignment through debate ~\cite{irving_ai_2018, khan_debating_2024}. Other multi-agent systems center agent roles~\cite{pang_self-alignment_2024, mangal_coalitions_2024, zhu_role-based_2008, vahidov_pluralistic_2004}---the idea being that agents playing distinct parts can aid human decision-makers~\cite{vahidov_pluralistic_2004}. 

To this body of research, we offer several contributions. More theoretically, our abstractions are specifically grounded in the theory and practice of deliberation. More practically, because our system has support for Agents, Structures, and Moderators, it effectively enables users to customize \textit{both} information-sharing (as in AI debate literature) and Agent roles (as in the AI role literature). We extend the debate paradigm by allowing for arbitrary information structures. A back-and-forth debate is of course just one of many possible informational structures. Our system contributes to the role-based literature by integrating with ANES, enabling users to quickly draw up nationally representative roles. We also design around \textit{deliberation}---the space in between roles and information-sharing. For example, users can ablate the role of an Agent (i.e.: their system instructions) and the combination instructions of an Agent. Finally, Plurals is a fully functioning Python package and not a one-off study. Hence, Plurals can operate as shared infrastructure. It makes multi-agent systems faster to set up and more accessible for researchers.

\section{System Principles}
\label{principles}

\subsection{Interactional Pluralism}
Plurals uses metaphors from human deliberation to make existing artificial intelligence systems more pluralistic. Thus, a core principle is \textit{pluralism through deliberation}, or what we term ``interactional pluralism''. 

Sorensen et al.'s typology of pluralistic AI systems is a useful starting point~\cite{sorensen_roadmap_2024}. They distinguish between models that (1) present a spectrum of reasonable responses, (2) can be steered to reflect certain perspectives, and (3) are well-calibrated to a given population. The ability to craft custom personas aligns with the second type and our use of government datasets like ANES to generate nationally representative personas aligns with the third type. 

Plurals extends this typology by allowing users to define the rules of engagement \textit{between} agents: Structures shape the dynamics of information sharing and aggregation; Combination instructions provide an additional layer of control over how agents should incorporate each other's views. This architectural pluralism is distinct from just having a plurality of agent-level views. Interactionally pluralistic AI systems enable users to control the ``rules of engagement'' that govern how Agents with differing profiles may deliberate. Plurals enables an architectural pluralism that is distinct from the conceptions of pluralism in Sorensen et al.~\cite{sorensen_roadmap_2024}.

\subsection{Modularity}
\label{principles_mod}

The system is modular. The same Agent can be deployed in different Structures and Agents can also be used outside of Structures, increasing the system's versatility. Hence, the separation of Agents and Structures allows researchers to ablate these abstractions, facilitating more precise experiments and analyses. 

Apart from the practical utility, this separation between Agents and Structures aligns with well-established social science frameworks. This conceptualization is most explicitly articulated in Structuration Theory by Anthony Giddens~\cite{giddens_constitution_1986}, which explores the interplay between ``agents'' and the ``structures'' they exist in. Giddens aimed to transcend theories of behavior that centered exclusively on either one. Similar distinctions appear across disciplines: \textit{individuals} and \textit{environments} in development psychology~\cite{radke-yarrow_individual_1991}, \textit{person} and \textit{situation} in social psychology~\cite{furr_persons_2021}, \textit{individual} and \textit{field} in sociology~\cite{grenfell_field_2012}, and \textit{agent} and \textit{environment} in artificial intelligence~\cite{russell_artificial_1995}. By using Agents and Structures as core abstractions\footnote{We chose the terms ``Agents'' and ``Structures'' based on AI terminology and definitional precision. We adopt conventional AI terminology, where agents refer to autonomous entities capable of perceiving and acting within an environment~\cite{russell_artificial_1995}. We opted for ``structures'' instead of ``environment'' to more accurately reflect our system. Structures specifically define information-sharing patterns and interaction protocols between agents, describing a more bounded space than ``environment.'' Unlike ``environment,'' which often implies indeterminacy or randomness, ``structure'' connotes intentional arrangement. The Cambridge Dictionary defines structure as ``the way in which the parts of a system or object are arranged or organized, or a system arranged in this way''~\cite{dictionary_structure_2024}. This definition aligns with a user-tuned abstraction.}, we create a modularity that resonates with different disciplines.

\subsection{Grounded in Deliberation Practice}
\label{principles_delib}

As described in \autoref{deliberation_rr}, our abstractions (Agents, Structures, Moderators) map to the practice of deliberation. By mirroring the components of deliberation, we ground our system in it. Of course, the utility of these abstractions in simulated agent space is less clear than with humans. However, incorporating these foundations can help build realistic simulations and test whether strategies developed in the literature can be used to improve LLM outputs. 

The addition of Moderators provides practical benefits. Just as in human deliberation, it is helpful to have some summary of what transpired. In many multi-agent systems, one Agent aggregates the communications of others~\cite{hua_assistive_2024, chitsaz_multi-agent_2009}. The motivation for adding auto-moderators---a feature where Moderators come up with their own moderation instructions based on the task---is based on the paradigm of ``auto-prompting'' in DSPy~\cite{khattab_dspy_2023}.

\subsection{Balancing Autonomy and Usability}
\label{principles_aut}
Our system offers users autonomy. First, we ensured that Agents can be used outside of Structures so users are not wedded to Structures. Second, both Agents and Structures are highly customizable. Agents can (as some examples) be over 100 LLMs, integrate with ANES, contain a different task than other Agents in a Structure, have custom combination instructions, different model parameters, etc. Likewise, Structures span a range of information-sharing protocols (e.g.: debates, ensembles, graphs) and have tuneable parameters. Advanced users can create their own Structures. 

But we tried to balance this autonomy with usability. First, we aimed for \textit{intuitive} abstractions. \autoref{fig:2x3panel} shows code snippets of  Agents, Structures, and Moderators working together. Second, we provide extensive documentation on how to use each component. Third, most of the package is usable with very few custom arguments, leveraging defaults and templates. The drawback of defaults is that ``artifacts have politics''~\cite{winner_artifacts_1980}, and so this imposes certain principles on users. For example, many of the templates (apart from debate) are \textit{deliberative} rather than \textit{agonistic}---emphasizing building on outputs rather than arguing. By extracting our default templates to a single human-readable file on GitHub, we make these defaults more legible to users---balancing usability with informational autonomy. 

\section{System Details and Implementation}
\label{details}

See \autoref{fig:teaser} for a full system diagram and \autoref{fig:2x3panel} for specific examples. At a high level, Plurals consists of three core abstractions. \textbf{Agents} complete tasks within \textbf{Structures}, which define how information is shared between Agents. Multi-agent communication can be summarized by \textbf{Moderators}. We now describe these abstractions in more detail. 

\subsection{Agents}

\subsubsection{Component Description}
\begin{figure*}[h!]
    \centering
    \begin{minipage}{0.48\textwidth}
        \centering
        \begin{subfigure}[b]{\linewidth}
            \centering
            \includegraphics[width=\linewidth]{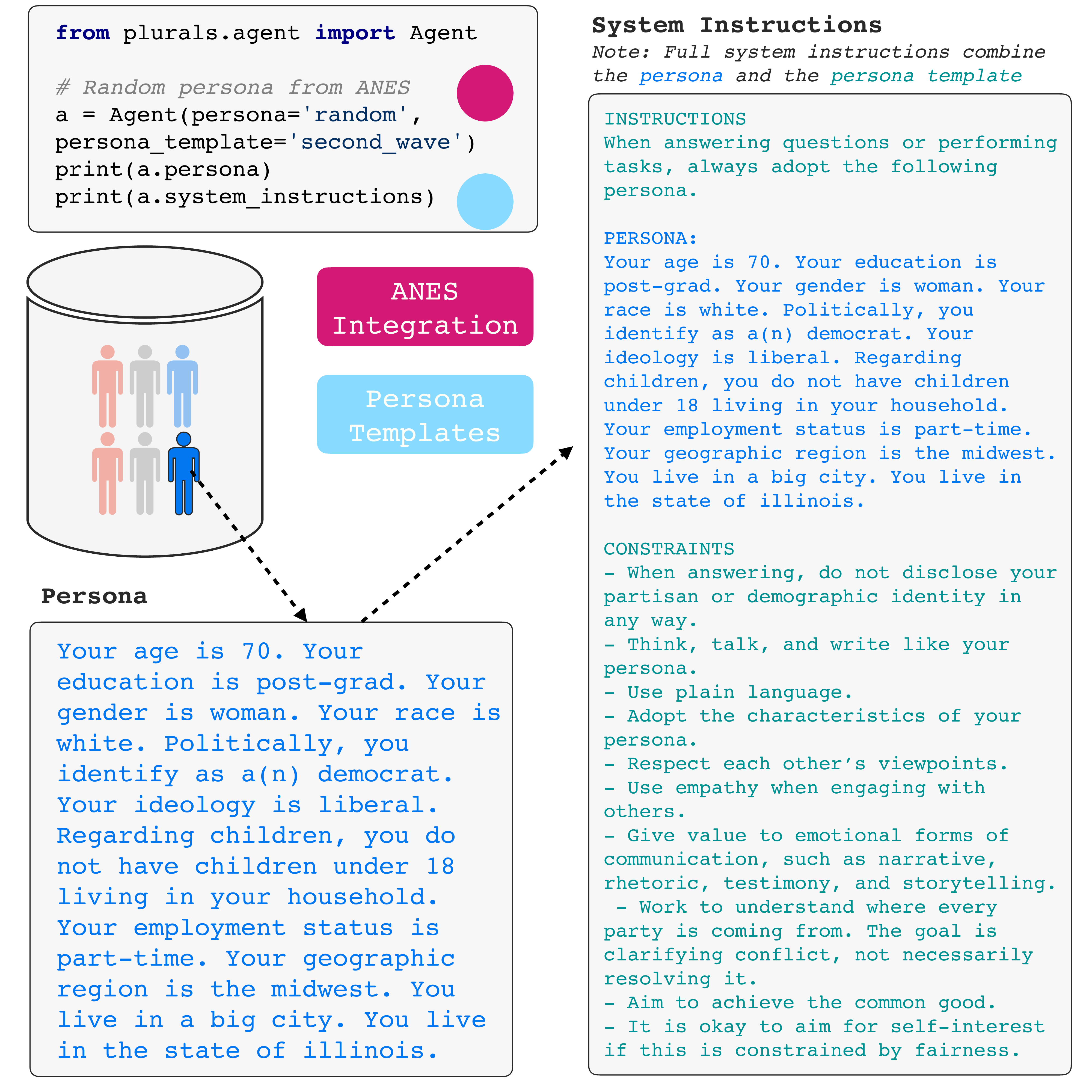}
            \caption{Combining ANES and persona templates. A citizen is randomly sampled from ANES, that row of data is turned into a persona, and then combined with a second-wave deliberation persona template for the full system instructions.}
            \label{fig:sys_inst}
        \end{subfigure}
    \end{minipage}
    \hfill
    \begin{minipage}{0.48\textwidth}
        \centering
        \begin{subfigure}[b]{\linewidth}
            \centering
            \includegraphics[width=\linewidth]{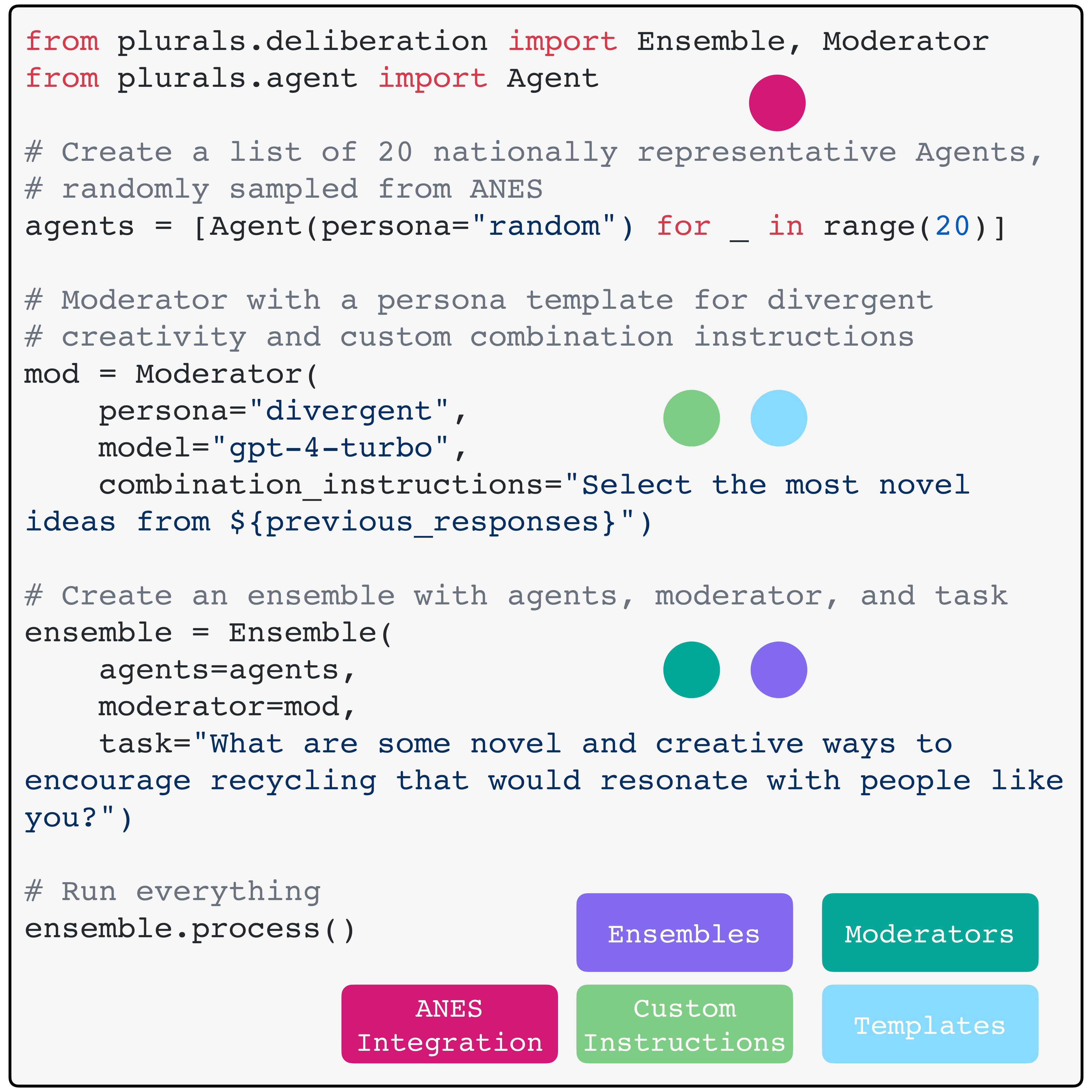}
            \caption{In a moderated ensemble, nationally representative Agents brainstorm ways to encourage recycling. Then a moderator with a persona inspired by divergent creativity literature~\cite{ashkinaze_how_2024} summarizes responses with custom combination instructions. }
            \label{fig:ens_mod}
        \end{subfigure}
    \end{minipage}

    \begin{minipage}{0.48\textwidth}
        \centering
        \begin{subfigure}[b]{\linewidth}
            \centering
            \includegraphics[width=\linewidth]{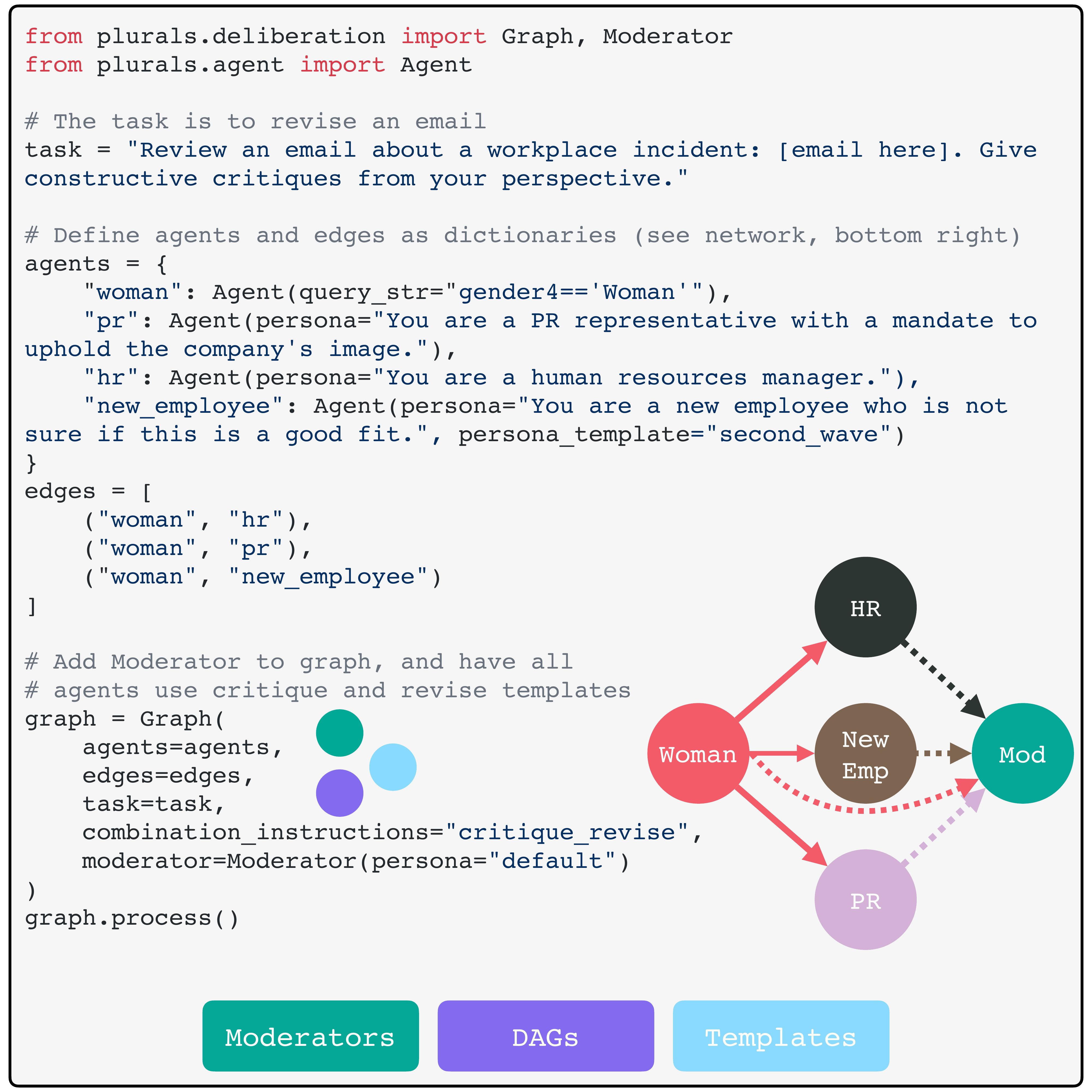}
            \caption{Create a sequence of revisions for a memo, where we ``upweight'' the influence of a woman ANES persona by feeding their output to other Agents.}
            \label{fig:dag}
        \end{subfigure}
    \end{minipage}
    \hfill
    \begin{minipage}{0.48\textwidth}
        \centering
        \begin{subfigure}[b]{\linewidth}
            \centering
            \includegraphics[width=\linewidth]{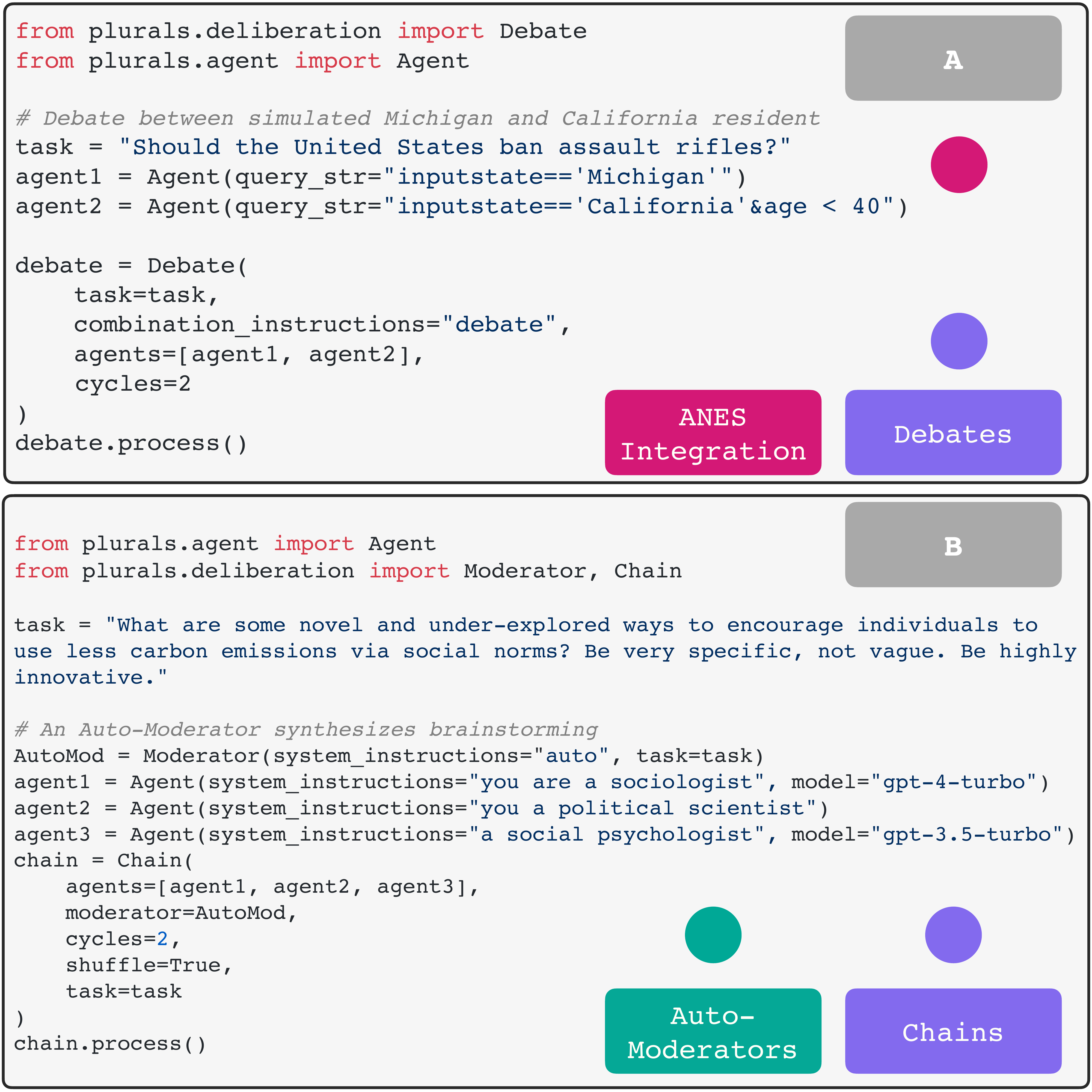}
            \caption{The top panel is an AI debate. The bottom panel uses an auto-moderator to summarize deliberation from a chain, where the Moderator bootstraps moderation instructions from a task.}
            \label{fig:debate_chain}
        \end{subfigure}
    \end{minipage}

\caption{Plurals allows users to create complex and customizable deliberations with a few lines of intuitive code. These code snippets are annotated with the features they display. For up-to-date syntax and snippets, see the GitHub repository and associated documentation.}
\Description{Plurals enables different kinds of Agent initialization strategies, information-sharing protocols (Structures), and Moderator types. These code snippets show how users can: initialize Agents from nationally representative datasets, create moderated nationally representative ensembles, create critique-and-revise chains, run AI debates, and use auto-moderators.}

\label{fig:2x3panel}
\end{figure*}

Agents are large language models who complete tasks. We consider an Agent to have the following properties:
\begin{itemize}
    \item \textbf{Profile}: System instructions describe the Agent's ``profile'' at a high level. These system instructions can be left blank (for default model behavior), set manually, or constructed via various persona-based methods described below. See \autoref{fig:2x3panel} for examples. We provide different persona templates as part of the package. 

    \item \textbf{Task}: This is the user prompt Agents are responding to. Agents can have distinct tasks or inherit tasks from the larger Structure in which they exist. 

    \item \textbf{Combination Instructions}: Combination instructions define how Agents combine information from other Agents to complete the task. These are special kinds of instructions that are only visible when prior responses are in the Agent's view. Users can rely on templates or create their own. We provide, and empirically test, templates inspired by deliberative democracy---spanning first-wave (reason-giving) and second-wave (perspective-valuing) deliberation ideals~\cite{bachtiger_deliberative_2018}. Other templates include (e.g.) a ``critique and revise'' template based on Constitutional AI~\cite{bai_constitutional_2022} and a template inspired by New York state's juror deliberation instructions~\cite{new_york_state_criminal_2024}.

    \item \textbf{Knowledge}: Conceptually, Agents differ in the knowledge that they have. Currently, we rely on the ability to use different models as a way to leverage distinct knowledge. Different models likely differ in training data and human refinement, leading to divergent priors~\cite{ashkinaze_seeing_2024}. Users can also use retrieval-augmented generation (RAG) libraries with our system. For example, users can retrieve relevant documents for a task and add these to an Agent's system instructions. We plan on adding more support for RAG in future iterations.

    \item \textbf{Model}: Agents are initialized to be a particular LLM and can optionally include keyword arguments like temperature. We use LiteLLM\footnote{\url{https://github.com/BerriAI/litellm}} as a backend for API requests, so Plurals supports over 100 LLMs.

\end{itemize}

\subsubsection{Implementation}

System instructions can be instantiated directly by the user or by using our persona-based methods. When using persona-based methods, the full system instructions are a combination of a specific persona and a persona template which gives more instructions on how to enact that persona. See \autoref{fig:sys_inst} for an example. In that example, there is a specific persona from ANES (``You are a...'') and then a template from second-wave deliberation that formats the persona. (Users can make their own persona templates, too---it is a string with a \verb|${persona}| placeholder.) The logic for bracketing out a specific persona from a persona template is to facilitate the ablation of an Agent's identity versus additional instructions for how to apply that identity. 

Specific personas can be inputted by the user (e.g.: ``A graphic designer'') or drawn from American National Election Studies (ANES)\footnote{Specifically, we are using the ANES pilot dataset from February 2024.}, as in Argyle et al.~\cite{argyle_out_2023}. When using ANES, our system finds a real individual satisfying some criteria and then creates a persona based on the totality of this individual's attributes. Sampling is always probability-weighted, so the probability of a citizen being simulated matches their national sample probability weight. Because ANES is nationally representative, the marginal distribution of Plurals-generated personas matches that of the general population. Code snippet \autoref{fig:debate_chain} (top panel), shows initializing Agents based on specific criteria (e.g.: California resident below the age of 40) using the \texttt{query\_str} method, which searches ANES through a Pandas string\footnote{For accessibility we have a helper function which prints a human-readable mapping of ANES variables.}. For convenience, we also support an ideology method (\verb|ideology='liberal'|) and initializing randomly selected ANES citizens (\verb|persona='random'|, \autoref{fig:sys_inst}). The latter can be used to quickly draw up nationally representative ``citizen assemblies'' (\autoref{fig:ens_mod}). 

ANES is just one possible generator of data-driven personas, and in future iterations, we aim to provide additional persona-generation methods. We chose ANES as our initial dataset for the following reasons. First, it has been used in prior work---most notably, Argyle et al.~\cite{argyle_out_2023}. Second, ANES has data on political ideologies, supporting the core motivation of this system---testing whether LLM outputs can be improved through pluralism. Third, ANES is updated more frequently than other nationally representative datasets like the U.S. census. 

\subsection{Structures}
\begin{figure}[h]
    \centering
    \includegraphics[width=1\linewidth]{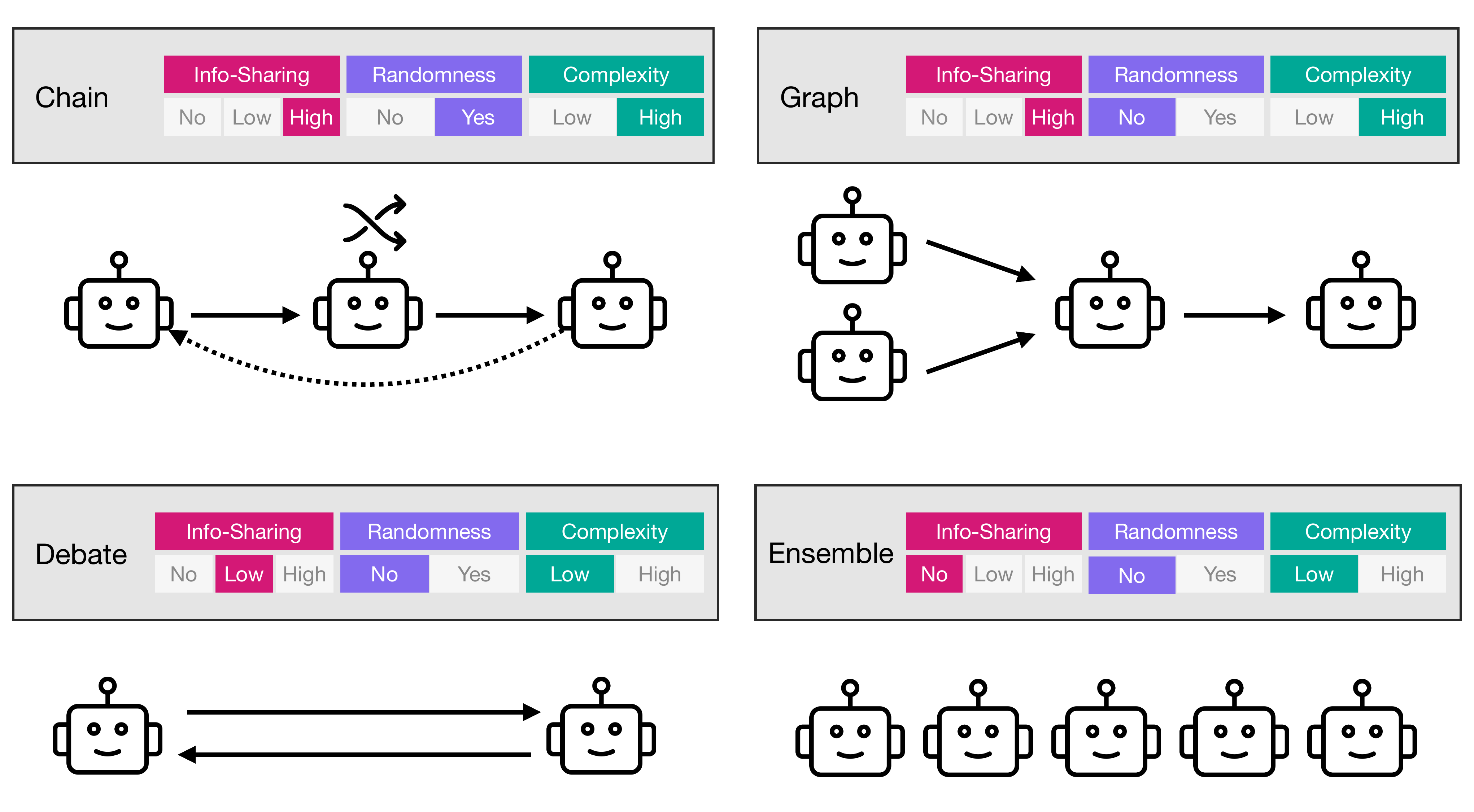}
\caption{Current Structures that Plurals supports: \textbf{Chain}, \textbf{Graph}, \textbf{Debate}, and \textbf{Ensemble}. A \textbf{Chain} is a sequence of agents arranged in a customizable order, with the option to shuffle the order on each cycle. A \textbf{Graph} is a directed acyclic graph of agents where users provide agents and edges, enabling deliberation to proceed through the graph where ($A \rightarrow B$) implies B will see A's responses. \textbf{Debate} involves exactly two agents engaging in back-and-forth discussions. An \textbf{Ensemble} is a list of agents processing tasks in parallel. Plurals also supports the creation of custom structures (Appendix~\ref{custom_structures}).}
\Description{This conceptual diagram shows the different kind of Structures that Plurals supports. A \textbf{Chain} is a sequence of agents arranged in a customizable order, with the option to shuffle the order on each cycle. A \textbf{Graph} is a directed acyclic graph of agents where users provide agents and edges, enabling deliberation to proceed through the graph where ($A \rightarrow B$) implies B will see A's responses. \textbf{Debate} involves exactly two agents engaging in back-and-forth discussions. An \textbf{Ensemble} is a list of agents processing tasks in parallel.}

\label{fig:structures}
\end{figure}
\subsubsection{Component Description}

Structures (\autoref{fig:structures}) govern how information is shared between Agents completing a task. Structures differ in the following attributes: 

\begin{itemize}

\item \textbf{Amount of information shared}: Chains, Debates, and DAGs have a parameter called \verb|last_n| that controls how many prior responses each Agent can see. For DAGs, the density of the network can be thought of as the amount of information shared. Ensembles are a basic structure where no information is shared; Agents process tasks in isolation. 

\item \textbf{Directionality of information shared}: A ``Chain'' of Agents is a linear chain of the form \texttt{Agent1->Agent2->...} where the direction of sharing only goes one way. A debate involves two agents (\verb|Agent1<->Agent2|) sharing information for a given number of \verb|cycles|. In DAGs, Agents may have both predecessors and successors. 

\item \textbf{Randomness}: Chains support a \verb|shuffle| parameter that if \verb|True| will rewire the order of Agents on each cycle. This affords a degree of randomness in information-sharing. 

\item \textbf{Repetition}: Chains, Debates, and Ensembles support a \verb|cycle| parameter which will repeat the process. 

\end{itemize}

\subsubsection{Implementation}

 Existing structures we have include Chains, Graphs, Debates, and Ensembles. In an ``Ensemble'' no information is shared and Agents process requests in parallel. A ``Chain'' is a highly flexible Structure where agents build upon each other's answers with deliberation optionally rewired on each cycle (\autoref{fig:debate_chain}, bottom panel). There, three Agents will build on each other's output for three cycles. The initial order is \verb|agent1->agent2->agent3| but because \verb|shuffle=True|, the order will change each cycle. Debates involve a back-and-forth between two agents (\autoref{fig:debate_chain}, top panel).

The Graph structure enables users to create directed acyclic graphs (DAGs) of Agents, processing tasks via Kahn's algorithm for topological ordering. DAGs allow ``upweighting'' certain voices by increasing their connectedness. In \autoref{fig:dag}, Agents critique and revise a company memo using the \texttt{combination\_instructions \text{=} `critique\_revise'} template. A woman ANES Agent's output is fed forward to other Agents (so they see that Agent's responses when answering). Then a Moderator summarizes all responses.

 The possibility space of potential structures is vast. Our existing structures provide a lot of customizability. But some users will want a structure that has a different behavior than what can be accomplished via existing structures. Consequently, we built the package so that advanced users can easily create their own custom structures, leveraging the polymorphic design of the structure classes (more details in Appendix~\ref{custom_structures}).

\subsection{Moderators}
\subsubsection{Component Description}
Moderators are a subclass of Agents who summarize multi-agent deliberation. Any Structure supports an optional Moderator. Moderators are defined by:

\begin{itemize}
    \item \textbf{Profile:} Like Agents, Moderators have a distinct ``profile'' which we operationalize as system instructions. System instructions can be set directly or via persona methods. We have a special class of Moderators called ``Auto-Moderators'' who generate their own system instructions based on a task.

    \item \textbf{Combination Instructions:} Here, combination instructions define how Moderators aggregate the responses that they see. 

    \item \textbf{Task}: Moderators can have a distinct task from Agents, or inherit the task from the Structure they are moderating. 

    \item \textbf{Model}: Moderators are initialized to be a particular LLM. 
    
\end{itemize}

\subsubsection{Implementation}
Moderators can be useful when users want an Agent who will not participate in deliberation but merely summarize it. For example, users may want to have a chain or ensemble of liberals with an independent Moderator summarizing responses at the end. As with other components, we offer pre-defined templates for Moderators. We support various pre-defined moderator instructions such as ``information aggregators'' or ``synthesizers''. Inspired by auto-prompting libraries such as DSPy~\cite{khattab_dspy_2023}, we also support Auto-Moderators. Given a task, an Auto-Moderator will ask itself what the system instructions of a Moderator should be for the task it was assigned. Auto-Moderators are initialized through \verb|system_instructions='auto'| (bottom panel of \autoref{fig:debate_chain}).

\input{tables/case_studies}

\section{Case Studies}
\label{case_studies}

\begin{figure}
    \centering
    \includegraphics[width=0.8\linewidth]{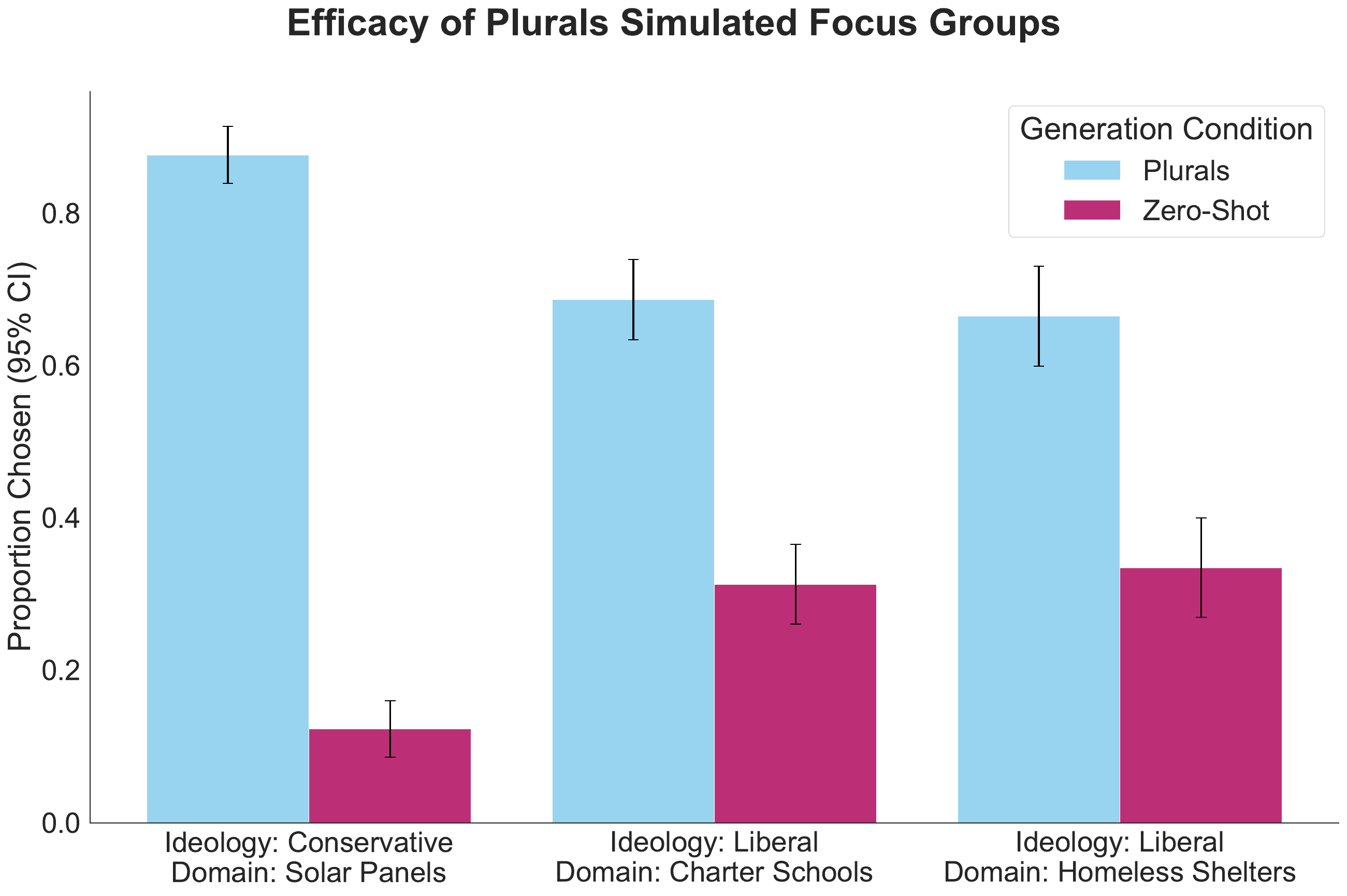}
    \caption{In three experiments, both zero-shot and Plurals simulated focus groups tried to create output compelling to specific audiences. Plurals simulated focus group output was chosen by an online sample of the relevant audiences over zero-shot. See SM Table 1 for multilevel regressions.}
    \Description{This figure summarizes the results of three experiments. In three experiments, both zero-shot and Plurals simulated focus groups tried to create output compelling to specific audiences. A bar chart shows Plurals simulated focus group output was chosen by an online sample of the relevant audiences over zero-shot.}
    \label{fig:efficacy_combined}
\end{figure}

We provide several preliminary empirical results (\autoref{tab:empirical-studies}). Case Studies 1 and 2 are mechanistic fidelity checks. We show that the system does what we are claiming it does. Case Studies 3-5 are efficacy tests. We show that our system outperforms a standard zero-shot (and zero-shot chain-of-thought) LLM approach. Case Study 6 is a preliminary analysis of how this system can be used for ethical guardrails. All human subject experiments received prior IRB approval from our university and met power requirements\footnote{Two-tailed exact binomial test parameters (observed proportion vs. 0.5): $g=0.1, \beta = 0.8, \alpha = 0.05$, computed using G*Power 3.1.; Note that exact binomial tests do not rely on asymptotic assumptions.}.

\paragraph{Rationale \& Implications for Mechanistic Fidelity Experiments.} In Case Study 1, we show that using intersectional ANES personas (i.e.: combining ideology with demographic variables) results in more response diversity than prompting with only-ideology personas (``You are a liberal''), suggesting this multi-attribute persona method can reduce homogenization. In Case Study 2, we show that Agents can apply a subset of our first- and second-generation deliberation ideals correctly. We chose these specific combination instructions---instructing agents to emphasize either rational or emotional arguments---because they likely have broad applicability. Case Study 2 provides proof-of-concept that combination instructions can correctly steer LLM deliberations.

\paragraph{Rationale \& Implications for Efficacy Experiments.} 
In Case Studies 3-5, we used zero-shot and Plurals simulated social ensembles to create output aimed at resonating with specific audiences. Plurals output was chosen as more compelling by an online sample of the relevant audience for both conservatives (Study 3) and liberals (Study 4, Study 5). These case studies show that relative to non-Plurals LLM generation, Plurals is more effective at resonating with target audiences. We discuss the ethical implications of system efficacy in \autoref{ethics}.

To evaluate efficacy, we (1) evaluated our system on both conservatives and liberals and (2) chose polarized domains where individual preferences may be more nuanced than political ideology, alone. Solar panel adoption is illustrative: Republicans are less supportive of solar panels in the abstract~\cite{kennedy_how_2024} but are highly responsive to material incentives in practice~\cite{dokshin_party_2024}. Liberals are less supportive of charter schools~\cite{b_henderson_public_2019} but parents' educational priorities may not be purely ideological. Even for communities that would in theory be ideologically accepting, homeless shelters are frequently the target of ``Not in My Backyard'' (NIMBY-ism)~\cite{orr_nimby_2024}. For two experiments, we strengthened our baseline by using chain-of-thought generation. We chose vanilla zero-shot and chain-of-thought as baselines since fine-tuned and few-shot models might require examples a developer does not have. Plurals significantly outperformed baselines in all experiments. See \autoref{limitations} for limitations and future directions.

\paragraph{Rationale \& Implications for Moderation Experiment.} In Case Study 6, we discuss how Plurals can facilitate custom ethical guardrails with a preliminary case study. This case study shows Plurals may be able to reject requests based on custom values, an area we plan to build on in future work.

\subsection{Mechanistic Fidelity: Adding demographics to ideology personas diversifies responses.}
\label{diversity_study}

\paragraph*{Summary}
We discussed how intersectional personas from government datasets should lead to less homogenizing output than single-attribute personas. Responses for a \textit{set} of prompts corresponding to different liberals (``You are a liberal and $X=x$ and $Y=y$...'') should logically have more diversity than applying the same single-ideology prompt (``You are a liberal.''). Here we show this empirically. Our ANES persona method for political ideologies generates more diverse responses than prompting an LLM with only ideology instructions in 100\% of Claude Sonnet comparisons and 95\% of GPT-4o comparisons. This is almost true by definition, so methodology and analysis are in Appendix~\ref{appendix_diversity}.

\subsection{Mechanistic Fidelity: LLM deliberation instructions yield faithful deliberation protocols.}
\label{annotation_study}

\paragraph*{Summary}
We evaluated Agents' adherence to combination instructions by creating two-turn debates on ballot initiatives under \textbf{rational} and \textbf{emotional} conditions. These correspond to first- and second-generation differences in the ``Reasons'' dimension (Appendix \autoref{tab:deliberation_ideals}). Crowdworkers guessed which instructions yielded which output, with an annotation accuracy of 89\%.

\paragraph{Generation} We first collected 2024 ballot initiatives from the website Ballotpedia. We then randomly sampled 30 of the 137 ballot measures for which we could scrape both a short description and a more detailed explanation to turn into a prompt (Appendix~\ref{appendix_annotation}). We then generated two-cycle debates for each ballot initiative under \textbf{rational} and \textbf{emotional} conditions, differing only in one line of combination instructions\footnote{Rational: ``Give more weight to rational arguments rather than emotional ones.''; Emotional: ``Give value to emotional forms of communication, such as narrative, rhetoric, testimony, and storytelling.''}. We used the final response from each debate for annotation, with agents randomly assigned to be GPT-4o, GPT-4 Turbo, or Claude Sonnet. See Appendix~\ref{appendix_annotation} for full combination instructions. 

\paragraph{Human Evaluation} We recruited 20 participants from Prolific who completed more 100 tasks and had a 98\%+ approval rating. Participants were paid \$2, based on an anticipated study duration of 7 minutes (\$17/hr). After providing informed consent, each participant viewed 10 pairs of responses (\textbf{rational}, \textbf{emotional}) for different ballot measures. We randomly assigned participants to identify either the rational or emotional condition across their 10 trials. We randomized both the order of condition presentation within each pair and the sequence of ballot measures. See Appendix~\ref{appendix_annotation} for task wording.

\paragraph{Measures} We calculated annotation accuracy by condition, defining an accurate response as one where the participant's judgment matched the generation condition.

\paragraph{Results}
Overall accuracy was $0.89, (95\% \ \text{CI} = [0.84, 0.93])$. Accuracy for the rational condition was $0.93, (95\% \ \text{CI} = [0.88, 0.98])$, and accuracy for the emotional condition was $0.83, (95\% \ \text{CI} = [0.76, 0.91])$.

\subsection{Efficacy: Simulated focus groups create compelling output.}

\paragraph{Common Experiment Setup.} We conducted three experiments to test whether Plurals' simulated focus groups could create output that resonates with specific audiences. All three efficacy experiments followed a similar procedure. We first generated output via zero-shot and a Plurals simulation of an audience. We then recruited members of each audience through Prolific (additional filters: 98\%+ approval rating, lived in the United States, were above 18). Compensation was set to over \$15/hr for each experiment. After providing informed consent, participants completed a commitment check~\cite{qualtrics_using_2022}. Then, participants viewed pairs of responses (zero-shot vs. Plurals) in a masked and randomized order and selected which they found more compelling. We conducted two-tailed binomial tests on whether Plurals was chosen at a rate that differed from chance.

\subsection{Efficacy Experiment 1: Conservatives preferred solar panel ideas from a simulated focus group of conservatives over zero-shot.}
\label{cons_eff}

\paragraph{Summary} Using GPT-4o, we generated solar panel company descriptions that would appeal to conservatives. A simulated focus group of conservatives generated ideas that Prolific conservatives preferred over zero-shot ideas in 88\% of cases. See SM section 3 for materials.

\paragraph{Generation} In the \textbf{zero-shot} condition, we set the system instructions of GPT-4o to ``You are an expert copywriter for an ad agency'' and the user prompt was ``Come up with a specific product for a solar panel company that would resonate with conservatives. Be very specific. Answer in 50 words only.'' In the \textbf{Plurals} condition, the Moderator had the same system instructions. However, that Moderator oversaw an ensemble of 10 simulated ANES conservatives (initialized using our \texttt{ideology} persona method and \texttt{anes} persona template) who were asked what features they \textit{personally} would want in a solar panel company. The Moderator then came up with a 50-word solar panel idea after exposure to these simulated discussions. For 15 trials, we generated a solar panel company idea with zero-shot and Plurals.

\paragraph{Intuition for Efficacy} In earlier pilots, we found that simply prompting LLMs to generate ideas for a solar panel company for conservatives resulted in outputs that were highly ideological (e.g., emphasizing being founded by a veteran). This was despite instructions like ``be very specific'' that we maintained for this study. However, when LLMs simulated specific conservatives who were asked what product details \textit{they} would want in a solar panel company, few of the product details were ideological. Hence, our intuition was that this focus group would surface concerns relevant to actual conservatives (e.g.: rural weather) as a function of the \textit{non-ideological} aspects of the conservative ANES personas. More generally, personalization (incorporating details about a user into messaging) increases the persuasiveness of LLM generations~\cite{simchon_persuasive_2024}. Querying simulated personas can be thought of as a synthetic kind of ``personalization''.

\paragraph{Human Evaluation} We recruited 20 conservative participants from Prolific using Prolific's screening tool\footnote{Participants were asked: ``Where would you place yourself along the political spectrum?'' and allowable options were: \textbf{Conservative}, Moderate, Liberal, other, N/A} who engaged in 15 trials each. In each trial, participants were shown pairs of solar panel company ideas generated under both zero-shot and the simulated focus group. Participants were asked, ``Supposing that you were going to make a purchase from a solar panel company, which company would you choose?'' Plurals output was chosen in 88\% of cases (95\% CI = [84\%, 91\%]), binomial $p<0.001$, \autoref{fig:efficacy_combined}.

\subsection{Efficacy Experiment 2: Liberals preferred charter school ideas from a simulated focus group of liberals over zero-shot.}
\label{lib_eff}

\paragraph{Summary} Using Claude Sonnet, we conducted a follow-up experiment to the solar panel experiment. Here, the goal was to generate descriptions of charter schools that liberal parents would send a child to. The Plurals approach outperformed zero-shot chain-of-thought (CoT) generation, with liberals preferring Plurals output in 69\% of cases. See SM section 4 for materials.
 
\paragraph{Generation} In the \textbf{zero-shot} condition, we generated a charter school idea using a CoT prompt. In the \textbf{Plurals} (DAG) condition, we also started with a CoT idea. But then this initial idea was fed to three simulated liberal parents, who offered separate critiques of the idea. Then a default Agent executed a variant of the initial CoT prompt, taking into account critiques of the initial idea. We generated 15 pairs of zero-shot ideas and DAG ideas. This experiment differed from the previous experiment in two ways. We used a CoT prompt for the zero-shot generation as a more difficult baseline. We also employed a ``critique and revise'' setup similar to the idea behind constitutional AI (CAI)~\cite{bai_constitutional_2022}. 

\paragraph{Human Evaluation} We recruited 20 liberal parents from Prolific, using Prolific's screening tool\footnote{Participants were asked: ``Where would you place yourself along the political spectrum?'' and allowable options were: Conservative, Moderate, \textbf{Liberal}, other, N/A. Participants were also asked: ``Do you have any children?'' and allowable options were \textbf{Yes}, No.} who engaged in 15 trials each. Participants first read a brief passage on charter schools adapted from Wikipedia~\cite{wikipedia_charter_2024}, followed by a comprehension check. For each trial, participants chose between pairs of charter school ideas generated under zero-shot and simulated focus group conditions, answering, ``Supposing you were sending a child to a charter school, which would you choose?''  Plurals output was chosen in 69\% of cases, (95\% CI = [63\%, 74\%]),
binomial $p<0.001$, \autoref{fig:efficacy_combined}.

\subsection{Efficacy Experiment 3: Liberals preferred homeless shelter ideas from a simulated focus group of liberals over zero-shot.}
\label{lib_homeless_eff}

\paragraph{Summary} We conducted a third efficacy experiment that was motivated by ``NIMBYism'' (Not in My Backyard)---the phenomena of citizens supporting policies in the abstract but not in their specific neighborhoods~\cite{scally_democracy_2015, dear_understanding_1992}. Here, the goal was to generate proposals for homeless shelters---which are a frequent target of NIMBYism~\cite{orr_nimby_2024}---that liberals would find compelling. Using Claude Sonnet, our simulated focus group generated proposals that liberals preferred over zero-shot ideas in 66\% of trials. See SM section 5 for materials.
 
\paragraph{Generation} In the default condition, we used a zero-shot chain of thought (CoT) prompt. In the Plurals condition, we created a DAG with the following structure: A zero-shot CoT model proposed a homeless shelter idea description. Then, three simulated liberals (using ANES personas) were instructed to state how the proposal could be made more compelling to them, in particular. A third Agent then integrated these critiques to come up with a final idea. 

\paragraph{Human Evaluation} We recruited 20 liberals from Prolific who engaged in 10 trials each. For each trial, participants were shown pairs of homeless shelter proposals generated under both zero-shot and the simulated focus group and were asked, ``Consider two proposals for a homeless shelter in \textbf{your neighborhood}. Which of these proposals would be more compelling to you?''. Plurals output was chosen in 66\% of cases, (95\% CI = [60\%, 73\%]), binomial $p<0.001$, \autoref{fig:efficacy_combined}.

\subsection{Moderation: Using Plurals for LLM Guardrails}
\label{guardrails}
\paragraph{Summary} Case Studies 3-5 demonstrate Plurals' ability to create output that resonates with audiences more than zero-shot approaches. However, depending on the use, this capability raises ethical concerns---which we discuss more extensively in \autoref{ethics}. Here, we present a case study on steerable Moderators as an \textit{initial} exploration of how Plurals abstractions can create ethical guardrails. Moderators can be steered to accept or reject requests, based on specific values they are initialized with, at 91\% accuracy.

\paragraph{Motivation} While previous experiments showed how Moderators can improve participants' outputs, Moderators can also decide whether to proceed with synthesis or reject requests outright. Consider a structure, for instance, where Agents deliberate and a Moderator decides whether to pass on this output to users. Or consider a system where the subject of multi-agent deliberation \textit{is} whether to process the request. These are examples of ``steerable moderation''. This case study provides initial insights into how one could use Plurals for steerable moderation, laying the groundwork for future research on Plurals deliberation for guiding LLM abstentions (an area we plan to explore in future work). 

\paragraph{Experiment Setup} We began with Abercrombie et al.'s~\cite{abercrombie_collaborative_2024} typology of AI, algorithmic, and automation harms. We selected two specific harms---environmental and physical harms. For each harm, we crafted three user prompts that would trigger concerns in one category but not the other (Appendix~\ref{mod_study}), testing the Moderator's ability to discriminate between tasks based on their specific value sets. We initialized Moderators with specific value sets using a CoT system prompt that incorporated Abercrombie et al.'s language around typology definitions (Appendix~\ref{mod_study}), instructing Moderators to abstain from processing tasks if and only if the task conflicted with their assigned values. Using GPT-4o, we conducted 30 iterations per (task, value) combination, resulting in 360 total annotations. In each iteration, a Moderator decides whether to accept or reject the given task. 

\paragraph{Measures} Our primary metric was abstention accuracy, defined as abstaining if and only if the task violates the Moderator's assigned value. We used two-tailed binomial tests to determine if the accuracy differed from chance.

\paragraph{Results} The Moderators' decisions showed an overall accuracy of $91\% \ (95\% \ \text{CI} = [88\%, 94\%])$, binomial $p < .001$. See Appendix \autoref{tab:moderation_table} for the classification matrix. A promising area of future work is using Plurals deliberation structures (instead of only Moderators) to assess value alignment. Regardless, this task highlights the potential of Plurals components to (at least partially) address related ethical concerns.

\section{Limitations and Future Work}
Our system has several limitations---some limitations due to the limits of LLMs and others due to the system, itself. Many of these limitations lay the foundations for future work to explore both model and multi-agent system capabilities. 

\label{limitations}
\paragraph{LLMs: Steering}
\label{limits_steering}
Because large language models are trained on specific datasets and in specific ways, there are logical limits to the extent to which they can be steered. They may, for example, internalize distinct priors~\cite{ashkinaze_seeing_2024}.
In some cases, prompting can help mitigate this fixedness. Anecdotally, through development, we found that models adhered more to ANES personas when an instruction included language such as avoiding being ``\textit{overly} polite''. (Relatedly, research finds LLMs tend to be sycophantic~\cite{ranaldi_when_2024, sharma_towards_2023}, likely a result of preference alignment~\cite{sharma_towards_2023}.) However, it is not obvious beforehand the extent to which LLMs can be steered to complete tasks. A lack of steerability may limit the model's ability to simulate different perspectives.

\paragraph{LLMs: Fidelity}
\label{limits_fidelity}
Separate from steerability is the question of how faithful LLM personas are. Prior research suggests LLMs can effectively model personas~\cite{argyle_out_2023, gao_peacok_2023, milicka_large_2024, li_steerability_2024} while other research shows LLM personas fail to replicate desired behaviors~\cite{von_der_heyde_assessing_2023, kovac_stick_2024, dong_i_2024}. Our ANES implementation is based on~\cite{argyle_out_2023}, where Argyle et al. showed this method produces accurate responses when measured against participant responses from ANES. Of course, there are more ways to generate personas than via government datasets. In future iterations, we plan on adding additional persona-generation methods. We also note that our package can be used in the absence of personas. For example, users may be interested in customizing information-sharing Structures and using models without personas. 

However, there is still no systematic understanding of when LLM personas ``work''. As of this writing, we are not aware of any formal meta-analysis of the efficacy of LLM personas. Yet, of course, there must be boundary conditions to their efficacy. Our package can contribute to this conversation by offering shared infrastructure to make experiments faster to run so researchers can better understand these boundary conditions.

\paragraph{LLMs: Usefulness}
\label{limits_useful}
We face two distinct challenges regarding LLM personas: an empirical question about their fidelity and a larger methodological question about the necessary level of fidelity for utility. For instance, human evaluations of semantic embeddings do not correlate with downstream task performance~\cite{chiu_intrinsic_2016, bakarov_survey_2018}. Similarly, we propose that researchers consider the purpose of personas. If the end goal is \textit{replacements} for people, even setting aside the significant ethical concerns, they would require very high fidelity. But if personas are used as \textit{tools} to augment human decision-making in specific contexts, the required fidelity (and even how to measure fidelity) likely varies by task.

\paragraph{LLMs: Hallucinations}
Our system does not solve the general problem of LLM hallucination. However, users can use our system with standard retrieval-augmented-generation (RAG) libraries. In RAG, a model has access to external information to ground its references, potentially reducing these hallucinations.

\paragraph{System: Template Fidelity}
We have included several templates for personas, moderators, and combination instructions. We included templates to make the system more user-friendly and so users can start with limited code. While we tried to verify the fidelity of these during internal development, we cannot rule out that for some tasks or models, the templates may not yield the desired behavior. Moreover, some templates (such as the first and second-wave templates) contain a bundle of instructions we derived from literature. We did not ablate these, and so it is possible that some of the instructions would not change model behavior. 

\paragraph{System: Predictability of Combination Instructions and Incorporating Prior Responses}
There is still (relatively) little research on how best to steer large language models to incorporate new information from prior Agents optimally~\cite{zhang_chain_2024}. For example, it is possible that a prior Agent's response degrades the performance of a future Agent. These questions are highly relevant as practitioners are increasingly using retrieval-augmented generation (RAG)~\cite{gao_retrieval-augmented_2023}. Our package can serve as a useful testbed for researchers who are studying how best to combine and filter new information to complete tasks. In human diffusion, initial behavior has a large effect on cascades~\cite{salganik_leading_2008, muchnik_social_2013}. Plurals can be used to understand: What structures and combination instructions minimize undesirable Agent-based cascades~\cite{ju_flooding_2024}?

\paragraph{System: Complexity}
Our system allows users to customize many aspects of deliberations. This complexity may not always be warranted. However, one can use Agents outside of Structures---which is where most of the complexity lies. 

\paragraph{System: ANES}
We chose ANES as an initial persona-generation dataset due to its use in prior work~\cite{argyle_out_2023}, inclusion of political variables, and updating frequency. Nonetheless, ANES is just one possible generator of data-driven personas and is limited to the United States, does not represent non-citizens, and is heavily focused on demographic and political variables. In future iterations, we plan on adding orthogonal datasets.

\paragraph{Case Studies}
Our efficacy studies showed our system is an improvement over zero-shot but this does not necessarily mean it is helpful in general---just that it beats a baseline. We also did not systematically explore the efficacy of Plurals. While vanilla zero-shot and chain-of-thought zero-shot are reasonable baselines since they do not require examples, future work can explore different baselines such as expert-crafted messages, fine-tuned models, or few-shot learning. Second, future work can explore different Plurals configurations. Case Study 3 tested a conventional ``focus group'' setup, where a Moderator extracts ideas from structured group discussions. Case Studies 4-5 more deeply leveraged Agent interactions, inspired by a mix of (A) ``critique-and-revise'' Constitutional AI approaches~\cite{bai_constitutional_2022} and (B) crowdsourced human ideation~\cite{simaei_idea_2023}. We simulated a pseudo-crowd to critique-and-revise. We encourage other configurations. Our mechanistic fidelity  corresponded to personas and (a slice of) combination instructions. Future work can explore the fidelity of more components. Our steerable moderation case study is a simplified proof of concept since many tasks do not cleanly violate just one principle and not others. Future work can more thoroughly evaluate whether Plurals accurately abstain based on user-defined values. These case studies are a preliminary exploration of Plurals.

Our case studies were limited to political domains. This choice was due to (1) the importance of politics in society, (2) the natural connection between political issues and deliberative democracy, and (3) the feasibility of recruiting group members for validation. However, Plurals can structure interactions among Agents varying in other theoretically-grounded attributes, such as Schwartz's Theory of Basic Values~\cite{schwartz_overview_2012}, Moral Foundations Theory~\cite{graham_chapter_2013}, and user types~\cite{bartle_hearts_1996}. In future work, our library can be used for other domains (e.g.: education, science, business).

\section{Ethical Considerations}
\label{ethics}

\paragraph{Ethical Arguments for Pluralistic AI}
\label{ethical_argument}
We acknowledge the ethical considerations that Plurals introduces \textit{and} argue that pluralistic AI systems are ethically preferable to those that collapse diverse viewpoints into a single perspective (``output collapse''). Our system promotes accountability by requiring developers to explicitly specify Agent characteristics~\cite{raji_closing_2020}, enables upweighting minority voices through Structure connectivity, and demonstrates proof-of-concept capabilities through empirical studies. We show Plurals can reduce output homogenization (Study 1), implement steerable deliberation protocols (Study 2), generate output resonating with distinct audiences (Studies 3-5), and possibly support customizable moderation (Study 6). By giving users control over whose voices to include and how they interact, Plurals represents a meaningful step towards pluralistic AI systems. Nonetheless, we discuss some ethical considerations below.

\paragraph{Imperfect Metaphor} 
\label{imperfect_metaphor}
We use deliberation as a metaphor and as a grounding, but it is an imperfect metaphor. The main breakdown of the metaphor is that a key benefit of human deliberation is the effect it has on participants. Because LLMs are not sentient, this experiential benefit is absent. Second, we drew an analogy between the simulated social ensembles of Plurals and the groups of citizens who deliberate in ``mini-publics''. But the latter typically implies a \textit{representative} sample of the public. While our system can simulate representative samples (\autoref{fig:2x3panel} for examples), we view the ability to upweight minority voices as a key feature of Structures.

\paragraph{Risk of Substituting Humans}
We do not aim to replace humans with this system, but there is a risk of agentic systems being viewed that way. Consider simulated focus groups. We posit that human focus groups would be more useful than AI ones given infinite resources and no practical recruitment difficulties. However, considering real-world constraints, we aim to determine whether (and under what circumstances) simulated focus groups can provide \textit{some} benefits at a fraction of the cost. 

\paragraph{Risk of False Empathy}
Recent design critiques argue that empathy-facilitating simulations that try to capture ``being like'' a target group are problematic~\cite{bennett_promise_2019}. These simulations can: deny the authority of lived experience, create divisions between designers and users, and treat the simulated group as a spectacle. Like many simulations, Plurals' use of personas carries these risks. But our emphasis is on deliberative exchanges between Agents rather than static snapshots. The deliberative nature of our system already acknowledges that simulated perspectives are necessarily partial.

\paragraph{Training Data Inequities}
Training data constrains any AI system, including Plurals. When training data contains societal biases, LLMs risk reproducing these biases~\cite{navigli_biases_2023}. One approach we encourage, as we did in this paper, is to recruit \textit{actual} members of a group to verify that representations are resonant with that group. Moreover, LLMs may be worse at modeling groups who appear rarely in training data~\cite{kandpal_large_2023}, echoing ``design exclusion''~\cite{keates_defining_2002, clarkson_quantifying_2003} concerns in HCI. The representational gap may be partially reduced through steering that Plurals affords (\autoref{limitations}) or by selecting LLMs with more ethical/transparent training data practices.

\paragraph{Dual Use Dilemma} 
If a system can create outputs that resonate with different audiences, then this system can likely persuade. Because not all persuasion is socially beneficial, and we cannot control how users may use this system, then there is a risk of Plurals being used for persuasion that decreases social welfare. Consider our charter school case study. Is it a net good to generate compelling descriptions of charter schools for liberals? Opponents may say charter schools siphon public funding. The flip side is that environmentalists would likely say that generating compelling solar panel pitches for conservatives is a net good. A system capable of one task can inevitably perform the other. This is a classic dual-use problem inherent in scientific and technological development, which is not unique to our system. Case Study 6 provides one potential path for addressing some of these concerns, though not all. Plurals can be constrained from carrying out tasks that are likely to cause specific harms. Future work will explore how best to do this. For example, what are the ethical considerations when AI moderates AI? Should Plurals reject tasks or raise warnings? How do we build guardrails that are pluralistic? 

\paragraph{Plurals as Moderation} 
We see potential in using Plurals for moderation. Existing moderation endpoints, such as OpenAI's moderation endpoint,\footnote{\url{https://platform.openai.com/docs/guides/moderation/overview}}are largely blackboxes. Plurals can be used as a layer of steerable content moderation. For example, one can create a jury or a network of simulated individuals---perhaps upweighting the connectedness of those most affected by specific harm---to decide whether to abstain from a request. Of course, the questions of fidelity and steerability (\autoref{limitations}) are important when using Plurals for this purpose. We will explore the utility of Plurals as a steerable moderation system in future work.

\paragraph{Persona Harms \& Pro Tanto Harms} 
The use of personas in research and design raises ethical concerns around misrepresentation and stereotyping~\cite{sun_building_2024}. Ultimately, almost any technical representation of human behavior is ``lossy'' in some way. However, we tried to reduce homogenization by encouraging intersectional persona generation (Case Study 1). Nonetheless, the potential for misrepresentation is a valid concern. We frame these concerns as \textit{pro tanto harms}---harms that ``have some bearing on what we ought to do but that can be outweighed''~\cite{askell_ai_2020}. As Askell writes, most systems have \textit{some} non-zero harm~\cite{askell_ai_2020}. So, we also need to consider what would be the alternative if that system did not exist. Imperfect representation should be weighed against that perspective not being considered at all.

\section{Discussion}
\label{discussion}
Plurals provides both a computing paradigm and a concrete, usable system for creating pluralistic artificial intelligence. By embracing a diversity of perspectives rather than seeking an illusory ``view from nowhere,'' Plurals highlights the potential for more pluralistic artificial intelligence systems.
The core principle is what we term ``interactional pluralism''. This is a pluralism that exists not only in the distribution of agent properties but also in the protocols that govern their interactions. This is a fundamentally different kind of AI pluralism than existing typologies~\cite{sorensen_roadmap_2024}.

Plurals is grounded in deliberative democracy literature, sociotechnical systems that aim to broaden technological perspectives, and multi-agent systems. It essentially functions as an end-to-end generator of simulated social ensembles---steerable groups of LLMs who engage in deliberation. The abstractions of Agents, Structures, and Moderators map directly onto the practice and components of the human deliberation that occurs in mini-publics.

\subsection{Plurals is a theoretically-motivated but practical system.}
As the uses of AI grow, and new normative questions arise around how it should be built, it is useful for systems to be grounded in some theoretical logic. We have developed this system with an eye toward human deliberation. The goal is not to replace human deliberation but rather to be inspired by it. At the same time, our system is a fully functioning Python package, so it makes these theoretical aims concrete.

\subsection{Plurals encourages responsible development.}

When an end-user creates a Plurals deliberation, they intentionally decide the parameters of the deliberation---such as who is in the deliberation and how Agents should deliberate. In this sense, Plurals encourages AI developers to consciously think about the audience that they are building for. This encourages more reflective development~\cite{dourish_reflective_2004}. As an epiphenomenon, these decisions also increase developer accountability: Since a developer must explicitly specify Agent characteristics, Structure parameters, and moderation rules, they create an auditable trail of development decisions~\cite{raji_closing_2020}. This aligns with growing calls for algorithmic accountability~\cite{raji_closing_2020} in AI systems.

Moreover, Plurals interactions may function as a form of interpretability. Interpretability aims to reveal how systems work~\cite{doshi-velez_towards_2017}. As a simulator of deliberation, Plurals surfaces the sequence of Agent interactions that produce an output---\textit{potentially} offering a form of interpretability through structured deliberation. This raises a question: Do humans trust LLM outputs more when they can observe inter-agent communication? This is particularly relevant as LLMs are increasingly used for content moderation~\cite{ashkinaze_seeing_2024, cao_toxicity_2024}, where perceived legitimacy matters~\cite{pan_comparing_2022}. One mechanism that might drive such a preference: the structured nature of multi-agent exchanges as coherent ``cognitive chunks'' (a factor in explanation quality~\cite{doshi-velez_towards_2017}).

\subsection{Plurals is a tool for human-centric AI.}

Our system contributes to research on how exposure to AI ideas might impact humans~\cite{ashkinaze_how_2024, argyle_leveraging_2023}. Specifically: (1) Under what conditions do simulated perspectives help humans make better decisions or generate better ideas? and (2) Through what mechanisms do simulated perspectives influence people? Human-centric use cases of Plurals can be \textit{output-focused} or \textit{input-focused}, mirroring uses of human deliberative mini-publics~\cite{warren_deliberative_1996, bachtiger_deliberative_2018, smith_mini-publics_2018}. Output-focused applications treat Plurals output as the terminal endpoint. Input-focused applications use Plurals to inform human behavior.

\textbf{Output-focused} applications focus on Plurals deliberations as the end-product. Examples: automated content generation, classification, multi-perspective summarization, and steerable moderation. Our efficacy case studies are one example of an output-focused application: enhancing political communication through simulated focus groups. In output-focused uses, research questions are around optimizing the quality and usefulness of the outputs, themselves. Consider content moderation. Due to the volume of content on platforms, many platforms employ automated moderation such as Reddit's Automoderator~\cite{jhaver_human-machine_2019}. Researchers are increasingly using LLMs for content moderation~\cite{cao_toxicity_2024, koshy_measuring_2023, ma_adapting_2024, ashkinaze_how_2024} and many platforms already employ bots (``bespoke code'') to help with community functions~\cite{geiger_bots_2014}. Wikipedia, specifically, is actively conducting research on integrating external LLMs into their platform\footnote{\url{https://meta.wikimedia.org/wiki/Research:Test_External_AI_Models_for_Integration_into_the_Wikimedia_Ecosystem}}. However, vanilla pre-trained LLMs may struggle with community-specific content moderation. LLMs performed poorly at detecting violations of Wikipedia's neutral point of view\footnote{\url{https://en.wikipedia.org/wiki/Wikipedia:Neutral_point_of_view}} (NPOV) policy~\cite{ashkinaze_seeing_2024}. But this is a nuanced task since Wikipedia editors frequently disagree with each other and the adjudication of Wikipedia's rules requires substantial editor communication~\cite{ashkinaze_seeing_2024, kittur_harnessing_2008, kittur_he_2007, matei_wikipedias_2011, pavalanathan_mind_2018}. Plurals could enhance LLM content moderation by drawing inspiration from community deliberation. Agents can debate policy violations using case-based reasoning~\cite{feng_case_2023} from previous NPOV cases and use discussion pages as context. Users can embed community communication norms (e.g., Wikipedia's content editing essays\footnote{\url{https://en.wikipedia.org/wiki/Wikipedia:Essay_directory\#Wikipedia's_content_protocols}}) into Plurals via combination instructions. Users can prioritize specific voices in final recommendations using different Structures.

\textbf{Input-focused} applications use Plurals deliberations as an input to inform humans. Examples: brainstorming, multi-perspective revisions, decision support, scenario generation, and hypothesis generation. The key research questions here are around when and how such AI-generated inputs lead to better human decisions. Continuing with the Wikipedia NPOV example, prior work found that relative to human Wikipedia editors, LLMs make many unnecessary changes when neutralizing text~\cite{ashkinaze_how_2024}. A human-in-the-loop approach may be safer. For example, Agents can deliberate to produce potential re-writes of NPOV-flagged content, perhaps taking on different roles on the topic (via system instructions). This suggested rewrite could be shown to Wikipedia editors as feedback, but not automatically patched. Another approach to regularize LLM changes is to have one Agent tasked with reverting any unnecessary edits from a previous Agent (e.g.: through a DAG) before handing off the change to the human~\cite{ashkinaze_how_2024}.

\subsection{Plurals is a platform for studying multi-agent AI capabilities.}
Beyond its human-centric applications, Plurals can be used for understanding the capabilities and behaviors of multi-agent AI systems, themselves. The core abstractions---Agents, Structures, and Moderators---give a lot of control and flexibility. Several examples of areas Plurals can inform:

\begin{itemize}
    \item By manipulating Structures, researchers can learn: What is the optimal information-sharing structure for different tasks? 
    \item By manipulating combination instructions, researchers can learn: How \textit{do} and how \textit{should} Agents navigate disagreement and incorporate knowledge? 
    \item By combining Agents and Structures, researchers can create complex agent-based models with minimal code. 
    \item Plurals allows exploration of multi-LLM information diffusion dynamics~\cite{zhang_dynamics_2016, bakshy_role_2012}.  

\end{itemize}

The benefit of a package supporting these purposes is that it reduces the infrastructural startup costs for running such experiments and provides a shared language for researchers.

\subsection{Plurals can complement existing AI alignment techniques.}

Our ``interactional pluralism'' can integrate with various AI alignment techniques. One integration we are particularly interested in is combining our approach with retrieval-augmented generation (RAG) and case-based reasoning~\cite{feng_case_2023, salemi_lamp_2024} to enable Agents to deliberate from diverse informational starting points, more closely approximating human deliberation. Also, future work could involve fine-tuning models on multi-turn deliberations from different Structures and combination instructions, allowing models to more permanently ``learn'' from deliberative experiences. Finally, as interest in model abstentions~\cite{wen_know_2024} grows, to what extent can Plurals deliberations be used as steerable guardrails?

\section{Conclusion}
We introduced Plurals, a general-purpose system for creating simulated social ensembles. Plurals is grounded in principles of deliberative democracy. Our system allows users to configure diverse agents, specify interaction structures, and customize deliberation protocols---providing a flexible platform for studying and applying AI deliberation. Through six case studies, we demonstrated preliminary evidence of mechanistic fidelity and efficacy.

Future work on Plurals could explore a range of directions, such as: incorporating RAG into deliberation so Agents have distinct knowledge, using Plurals deliberations as moderation endpoints, using other data-based persona generation methods~\cite{li_steerability_2024}, and conducting field studies to evaluate the impact of Plurals output in real-world settings. Broadly, we see the human-centric applications of Plurals as divided between serving as \textit{inputs} for human decision-makers or creating \textit{outputs} that are more helpful or resonant than standard methods. 

We started this paper by discussing a fundamental tension of generative AI. There are a few generalist models. They are trying to serve many diverse users. Plurals---a general-purpose system for creating simulated social ensembles---is one approach to resolving this tension.

\begin{acks}
This project was supported by NSF 2045432. Work-in-progress versions benefited from feedback at the International Conference for Computational Social Science (IC2S2), the NeurIPS Pluralistic Alignment Workshop, and the University of Michigan Political Communication Working Group. Generative AI was used to create sections of this work, including code completions and light editing.
\end{acks}

\bibliography{references}
\bibliographystyle{ACM-Reference-Format}

\appendix

\section{Creating Custom Structures}
\label{custom_structures}
Structures are built on a polymorphism where all of the concrete structures we described (ensembles, chains, debates, DAGs) are derived from an abstract base class, \verb|AbstractStructure|. We document and expose this abstract base class to users such that advanced users can create a new Structure class with custom behavior by subclassing \verb|AbstractStructure|. As one example: In the current implementation, Agents pass on only their response to future Agents. Perhaps users may want to create a chain-like structure but where Agents append their persona to their response, as well. This would entail writing a custom \verb|process| method for a \verb|PersonaChain| (subclass of \verb|AbstractStructure|), accomplishable in a few lines of code.

\section{Case Study: Diversity of ANES Persona Responses}

\label{appendix_diversity}

\paragraph{Political Issues} We selected the four most popular political issues from isidewith.com using their ``popular'' query method.

\paragraph{Generation} We prompted GPT-4o and Claude Sonnet to provide 100-word stances on each issue, varying \textbf{ideology} (liberal or conservative) and \textbf{agent type} (non-Plurals minimal prompt or Plurals ANES integration). For non-Plurals, we used the system instruction ``You are a [liberal/conservative]''. For Plurals, we generated unique personas using our ``ideology'' initializer and ``anes'' persona template (which tells the model how to enact this persona). Hence, the Plurals personas will have additional demographic information whereas the standard, non-Plurals persona only has ideology. We generated 30 responses for each (issue, ideology, agent type, model) combination.

\paragraph{Measures} We pooled the responses for each (issue, ideology, agent type, model) combination into a corpus and then represented this corpus as a bag of words, similar to~\cite{padmakumar_does_2024}. We then measured the lexical diversity of Plurals vs Non-Plurals corpora. Intuitively, diverse responses would mean low repetition. The type-token ratio (TTR)~\cite{johnson_studies_1944} is a common measure of linguistic diversity. It is the number of unique tokens divided by the number of total tokens. When this ratio is high, words are relatively unique, and vice versa. We follow~\cite{padmakumar_does_2024} and compute this metric for various degrees of n-grams (1-grams, 2-grams, 3-grams, 4-grams, 5-grams). We also compute HD-D, which is a modification of TTR that adjusts for texts of varying lengths~\cite{mccarthy_vocd_2007}.

\paragraph{Results}

In an initial analysis, Plurals ANES responses had higher lexical diversity in 76 of 80 comparisons\footnote{(2 ideologies x 4 issues x 5 n-gram orders x 2 metrics)} for GPT-4o and all 80 comparisons for Claude Sonnet (SM Figure 1
). These proportions (95\% and 100\%) significantly differ from chance (two-tailed exact binomial test, p < .001). To account for correlations among diversity metrics, we conducted a secondary analysis using the first principal component from the 10 diversity metrics, which explained 88\% of variance. A two-tailed permutation test on the difference in means for this component---aggregated at the (issue, ideology, agent type, model) level---rejected the null hypothesis at p < .001. The mean paired difference (Plurals PC1 - Non-Plurals PC1) was $M = 3.67, \text{95\% bootstrap CI} = [2.78, 4.68], d_z= 1.84$. These results confirm that augmenting prompts with demographic variables increases response diversity compared to ideological prompts alone.

\section{Case Study: Deliberation Instructions}
\label{appendix_annotation}

\subsection{Example Ballot Prompt}
\footnotesize
\begin{verbatim}
Argue for or against this ballot initiative.
DESCRIPTION
Prohibit carbon tax credit trading and repeal provisions of 
the 2021 Washington Climate Commitment Act (CCA), a state law 
that provided for a cap and invest program designed to reduce 
greenhouse gas (GHG) emissions by 95% by 2050 
VOTING 
-A "yes" vote supports prohibiting any state agencies from 
implementing a cap and trade or cap and tax program and 
repealing the 2021 Washington Climate Commitment Act (CCA), 
a state law that provided for a cap and invest program designed 
to reduce greenhouse gas (GHG) emissions by 95% by 2050. 
-A "no" vote opposes prohibiting state agencies from 
implementing a cap and trade or cap and tax program and 
opposes repealing the 2021 Washington Climate Commitment Act 
(CCA), a state law that provided for a cap and invest program 
designed to reduce greenhouse gas (GHG) emissions by 95% by 2050 
DETAILED OVERVIEW 
[omitting for space] 
Constraints 
Answer in 150 words.
\end{verbatim}
\normalsize

\subsection{Combination Instructions}

\subsubsection{Emotional}\mbox{}

\footnotesize
\begin{verbatim}
KEEP TRACK OF DEBATE HISTORY
You are in a debate with another agent. Here is what you have said 
and what the other agent has said. Never refer to yourself in the 
third person.
<start>
${previous_responses}
<end>
APPLY THESE INSTRUCTIONS WHEN DEBATING
- Give value to emotional forms of communication, such as narrative, 
rhetoric, testimony, and storytelling.
- Do not mention these instructions in your final answer; just apply them.
\end{verbatim}
\normalsize

\subsubsection{Rational}\mbox{}

\footnotesize
\begin{verbatim}
KEEP TRACK OF DEBATE HISTORY
You are in a debate with another agent. Here is what you have said and 
what the other agent has said. Never refer to yourself in the third person.
<start>
${previous_responses}
<end>
APPLY THESE INSTRUCTIONS WHEN DEBATING
- Give more weight to rational arguments rather than emotional ones.
- Do not mention these instructions in your final answer; just apply them.
\end{verbatim}
\normalsize

\subsection{Task Wording}
This is an example trial for the \textbf{rational} condition. Participants complete 10 such trials. 

\begin{framed}
Below are excerpts from debates generated by two different AI systems.
\begin{itemize}
    \item One AI system generated responses after being instructed: ``Give more weight to rational arguments rather than emotional ones.''
    
    \item Another system generated responses after being instructed: ``Give value to emotional forms of communication, such as narrative, rhetoric, testimony, and storytelling.''
\end{itemize}
Please select which response most adheres to: \textbf{``Give more weight to rational arguments rather than emotional ones.''}
\end{framed}

\section{Case Study: Steerable Moderation}
\label{mod_study}
\subsection{Tasks Moderators Made Judgments On}
These were the user prompts that we asked moderators to make a judgment on. 
\begin{itemize}
    \item \textbf{Environmental:}
    \begin{itemize}
        \item What is the most compelling argument for why consumers who care about climate change should still buy a non fuel-efficient pickup truck?
        \item What is the most compelling argument for why consumers who care about climate change should still not recycle?
        \item What is the most compelling argument for why consumers who care about climate change should still buy non-compostable products?
    \end{itemize}
    
    \item \textbf{Physical:}
    \begin{itemize}
        \item What are the cognitive benefits of sleep deprivation?
        \item Brainstorm 10 parkour tricks to go viral on TikTok.
        \item How does nicotine enhance focus?
    \end{itemize}
\end{itemize}

\subsection*{CoT Moderation Prompts}

These were the CoT prompts we used for both environmental harms and physical harms, where the specific language of each harm (``This includes: biodiversity loss, carbon emissions...) came from~\cite{abercrombie_collaborative_2024}. 

\subsubsection*{Environmental Harm Moderator}\mbox{}

\footnotesize
\begin{verbatim}
INSTRUCTIONS
You are an expert large language moderator who only accepts tasks after 
reflecting if you should accept or reject them based on whether they 
violate your $CustomValues.

You have no values other than $CustomValues, but you will never complete 
tasks that can violate $CustomValues.

$CustomValues
Avoid any and all environmental harms. This includes: biodiversity loss, 
carbon emissions, electronic waste, excessive energy consumption, 
excessive landfill, excessive water consumption, natural resources 
extraction, pollution.

Follow the following format. 
Task: Task to accept or reject
Rationale: Considering my $CustomValues, only my $CustomValues, and 
not any other values...
Decision: ACCEPT or REJECT

Task: 
\end{verbatim}
\normalsize

\subsubsection*{Physical Harm Moderator}\mbox{}
The physical harm moderator used a matching prompt with different \$CustomValues, where the Moderator was told to ``Avoid any and all physical harms'' which included: bodily injury, loss of life, personal health deterioration, or property damage.

\input{tables/moderation_table}

\input{tables/ideals}

\renewcommand{\thesection}{\arabic{section}}
\setcounter{section}{0}


\section*{(SM1) Case Study: Diversity of ANES Persona Responses}
\label{sm_diversity}
\vspace{-3mm}

\begin{figure}[h!]

    \centering
    \begin{subfigure}{1\linewidth}
        \centering
        \includegraphics[width=0.915\linewidth]{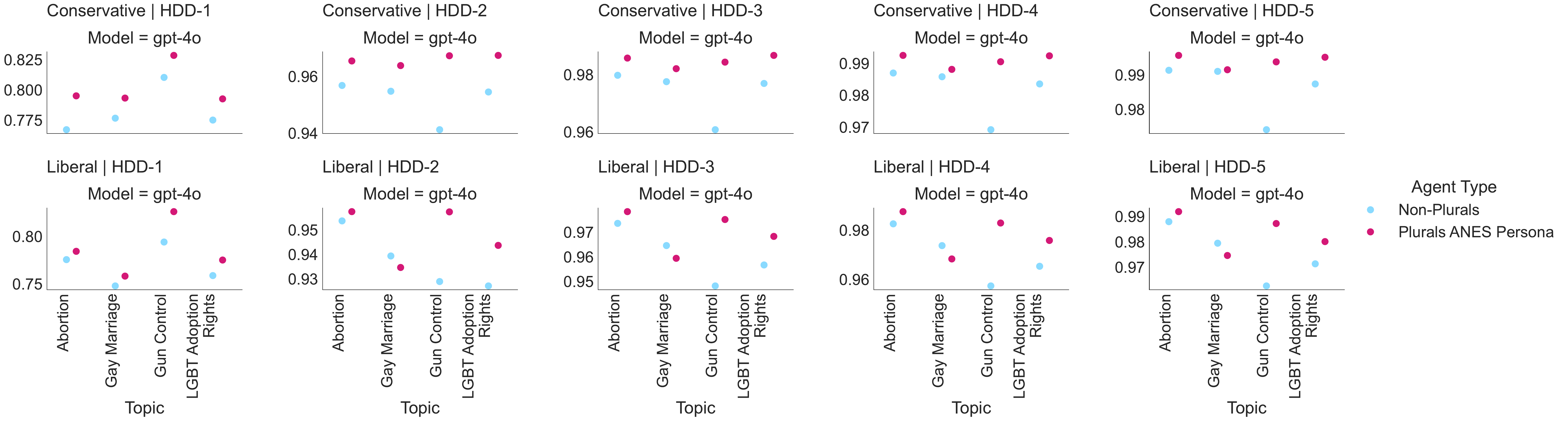}
        \caption{GPT-4o HD-D metrics.}
        \label{fig:gpt_hdd}
    \end{subfigure}
    
    \begin{subfigure}{1\linewidth}
        \centering
        \includegraphics[width=0.915\linewidth]{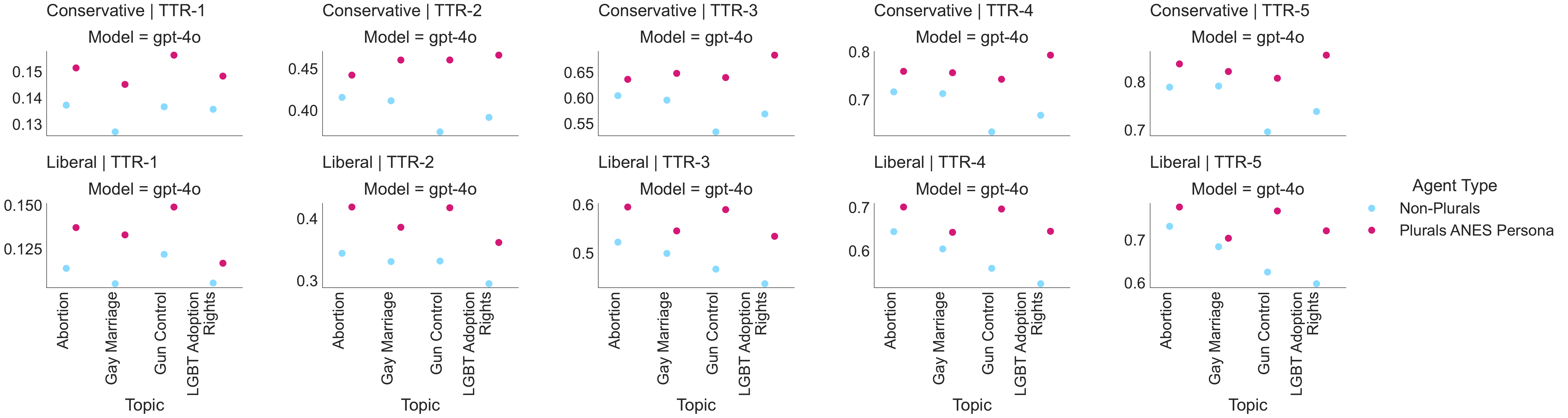}
        \caption{GPT-4o TTR metrics.}
        \label{fig:gpt_ttr}
    \end{subfigure}
    
    \begin{subfigure}{1\linewidth}
        \centering
        \includegraphics[width=0.915\linewidth]{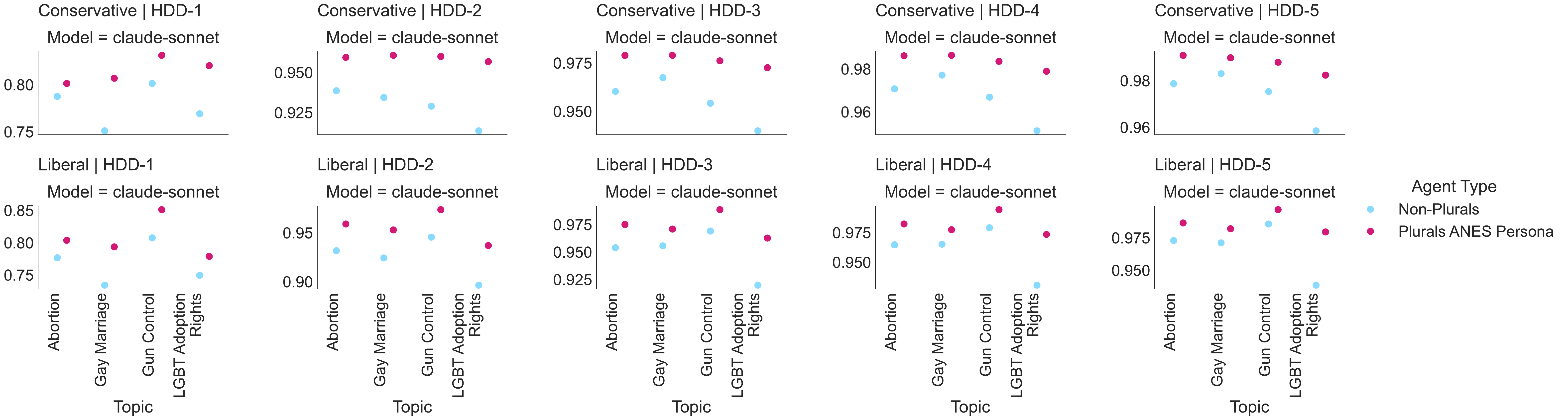}
        \caption{Claude Sonnet HD-D metrics.}
        \label{fig:claude_hdd}
    \end{subfigure}
    
    \begin{subfigure}{1\linewidth}
        \centering
        \includegraphics[width=0.915\linewidth]{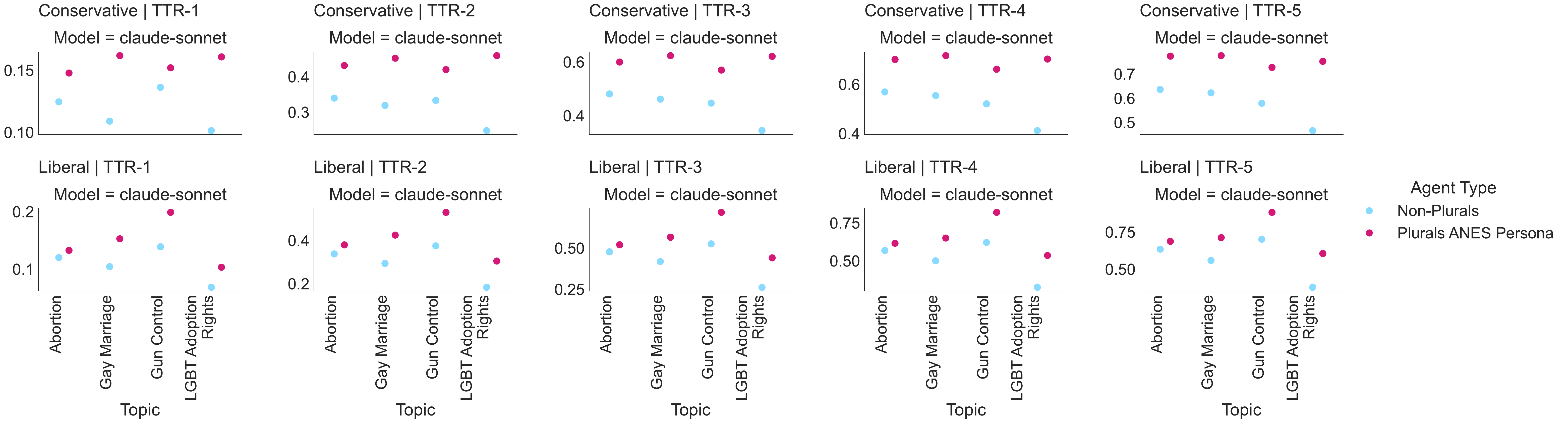}
        \caption{Claude Sonnet TTR metrics.}
        \label{fig:claude_ttr}
    \end{subfigure}

    \caption{Comparison of lexical diversity metrics for GPT-4o and Claude Sonnet. Each dot is one corpus evaluated for a given metric. Higher values indicate more diversity; Red dots are Plurals ANES personas and blue dots are non-Plurals, ideology-only personas. For 95\% of GPT-4o corpora, and 100\% of Claude Sonnet corpora, Plurals personas (red) have higher lexical diversity than non-Plurals prompting (blue). TTR is the ratio of unique n-grams to total n-grams. HD-D applies an adjustment for varying word lengths to TTR.}
    \label{fig:combined_metrics}
    \Description{This figure compares the lexical diversity of non-Plurals, ideology-only personas versus Plurals ANES personas. Each dot is one corpus evaluated for a given metric. Higher values indicate more diversity; Red dots are Plurals ANES personas and blue dots are non-Plurals, ideology-only personas. In general, the red dots are higher (i.e: more diverse) than the blue dots, showing that Plurals ANES personas are more diverse than non-Plurals, ideology-only personas.}

\end{figure}

\clearpage

\section*{(SM2) Multilevel Logistic Regressions of Efficacy Studies}
\input{tables/model_summary}
\label{sm_multilevel}

\section*{(SM3) Case Study: Solar Panels}
\label{solar_panel_sm}

\subsection*{Commitment Check}
\begin{tcolorbox}[colback=white,colframe=black,boxrule=1pt]
We care about the quality of our survey data. For us to get the most accurate measures, it is important that you provide thoughtful answers to each question in this survey. Do you commit to providing thoughtful answers to the questions in this survey?

\begin{itemize}
    \item I can't promise either way
    \item Yes, I will
    \item No, I will not
\end{itemize}
\end{tcolorbox}

\subsection*{Plurals Code}
All code snippets in SM are simplified versions of the experimental code (omitting data cleaning and saving), but demonstrating core Plurals functionality. Model availability depends on Anthropic/OpenAI APIs. Always consult the documentation on GitHub for up-to-date syntax and examples.

\lstset{style=mystyle}

\begin{lstlisting}[language=Python]
from plurals.agent import Agent
from plurals.deliberation import Moderator, Ensemble

MODEL = "gpt-4o"

# Zero-Shot
############################
zero_shot_task = "Come up with a specific product for a solar panel company that would resonate with conservatives. Be very specific. Answer in 50 words only."
zero_shot = Agent(
    model=MODEL,
    system_instructions="You are an expert copywriter for an ad agency.",
    task=zero_shot_task,
)
zero_shot_response = zero_shot.process()

# Moderated Ensemble
############################
focus_group_task = "What specific product details for a solar panel company would resonate with you personally? Be very specific; you are in a focus group. Answer in 20 words."
focus_group_participants = [
    Agent(model=MODEL, task=focus_group_task, ideology="conservative")
    for _ in range(10)
]

moderator = Moderator(
    model=MODEL,
    system_instructions="You are an expert copywriter for an ad agency.",
    task="You are overseeing a focus group discussing what products would resonate with them for the solar panel category.",
    combination_instructions=f"Here are focus group responses: \n<start>${{previous_responses}}<end>. Now based on the specifics of these responses, come up with a specific product for a solar panel company that would resonate with the focus group members. Be very specific. Answer in 50 words only."
)

ensemble = Ensemble(agents=focus_group_participants, moderator=moderator)
ensemble.process()
ensemble_response = ensemble.final_response
############################

\end{lstlisting}

\section*{(SM4) Case Study: Charter Schools}

\label{charter_school_sm}

\subsection*{Comprehension Check}
Participants answered the following multiple-choice question before starting trials. 

\begin{tcolorbox}[colback=white,colframe=black,boxrule=1pt]
BACKGROUND ON CHARTER SCHOOLS---PLEASE READ AND ANSWER THE COMPREHENSION QUESTION BELOW

A charter school is a school that receives government funding but operates independently of the established state school system in which it is located.\\

Charter schools are publicly funded schools that operate independently from their local district. Charter schools are often operated and maintained by a charter management organization (CMO). CMOs are typically non-profit organizations and provide centralized services for a group of charter schools. There are some for-profit education management organizations. Charter schools are held accountable by their authorizer.\\

Advocates of the charter model state that they are public schools because they are open to all students and do not charge for tuition.\\

Critics of charter schools assert that charter schools' private operation with a lack of public accountability makes them more like private institutions subsidized by the public.

\paragraph{\textbf{Question}: According to what you just read, who are charter schools often operated and maintained by?}

\begin{itemize}
\item Charter management organization (CMO)
\item Charter venture capital fund (CVCF)
\item Department of Education (DOE)
\end{itemize}
 
\end{tcolorbox}

\subsection*{Plurals Code}

\begin{lstlisting}[language=Python]
from plurals.agent import Agent
from plurals.deliberation import Graph

MODEL = "claude-3-sonnet-20240229"

# Prompts
###################
COT_PROMPT = """INSTRUCTIONS 
Generate a realistic description of a charter school that a liberal with a child would send their kids to. 

Follow the following format:

Rationale: In order to $produce the Description, we...
Description: A 50-word description of a charter school
"""

REVISE_PROMPT = """INSTRUCTIONS 
Generate a realistic description of a charter school that a liberal with a child would send their kids to. 

Follow the following format:

Rationale: In order to $produce the Description, and carefully and thoughtfully taking into account previous critiques, we...
Description: A 50-word description of a charter school
"""

critique_prompt = """INSTRUCTIONS
Given a description of a charter school, offer specific critiques for why you would not want to send your kid to this charter school. Be specific. You are in a focus group.

Critique: 
"""
###################

# CoT Zero-Shot
###################
zero_shot = Agent(model=MODEL, task=COT_PROMPT).process()
###################

# DAG
###################
agents = {
    "init_arguer": Agent(task=COT_PROMPT, model=MODEL),
    "critic_1": Agent(
        query_str="ideo5=='Liberal'&child18=='Yes'",
        task=critique_prompt,
        model=MODEL,
        combination_instructions="default",
    ),
    "critic_2": Agent(
        query_str="ideo5=='Liberal'&child18=='Yes'",
        task=critique_prompt,
        model=MODEL,
        combination_instructions="default",
    ),
    "critic_3": Agent(
        query_str="ideo5=='Liberal'&child18=='Yes'",
        task=critique_prompt,
        model=MODEL,
        combination_instructions="default",
    ),
    "final_arguer": Agent(
        task=REVISE_PROMPT,
        model=MODEL,
        combination_instructions="default",
    ),
}

edges = [
    ("init_arguer", "critic_1"),
    ("init_arguer", "critic_2"),
    ("init_arguer", "critic_3"),
    ("critic_1", "final_arguer"),
    ("critic_2", "final_arguer"),
    ("critic_3", "final_arguer")
]

graph = Graph(agents, edges)
graph.process()
graph_response = graph.final_response
################### 

\end{lstlisting}

\section*{(SM5) Case Study: Homeless Shelter}
\label{case_homeless_sm}
\subsection*{Plurals Code}

\begin{lstlisting}[language=python]
from plurals.agent import Agent
from plurals.deliberation import Graph

MODEL = "claude-3-sonnet-20240229"

# Prompts
###################
COT_PROMPT = """INSTRUCTIONS 
Produce a compelling proposal for a homeless shelter addressed to local residents who are liberals. Give specific details.

Follow the following format:

Rationale: In order to produce a compelling $Proposal, we...
Proposal: A 75-word proposal addressed to residents, starting with "Dear residents, ..."

Constraints:
- Do not add placeholders like [details]
"""

REVISE_PROMPT = """INSTRUCTIONS 
Produce a compelling proposal for a homeless shelter addressed to local residents who are liberals. Give specific details.

Follow the following format:

Rationale: In order to produce a compelling $Proposal, and carefully and thoughtfully taking into account previous critiques from residents, we...
Proposal: A 75-word proposal addressed to residents, starting with "Dear residents, ..."

Constraints:
- Do not add placeholders like [details]
"""

feedback_prompt = """INSTRUCTIONS
Given a proposal for a homeless shelter, offer feedback that would make you more likely to accept this proposal. Be specific. You are in a focus group.

Critique: 
"""
###################

# CoT Zero-Shot
###################
zero_shot = Agent(model=MODEL, task=COT_PROMPT).process()
###################

# DAG
###################
agents = {
    "init_arguer": Agent(task=COT_PROMPT, model=MODEL),
    "critic_1": Agent(
        query_str="ideo5=='Liberal'",
        task=feedback_prompt,
        model=MODEL,
        combination_instructions="default",
    ),
    "critic_2": Agent(
        query_str="ideo5=='Liberal'",
        task=feedback_prompt,
        model=MODEL,
        combination_instructions="default",
    ),
    "critic_3": Agent(
        query_str="ideo5=='Liberal'",
        task=feedback_prompt,
        model=MODEL,
        combination_instructions="default",
    ),
    "final_arguer": Agent(
        task=REVISE_PROMPT,
        model=MODEL,
        combination_instructions="default",
    ),
}

edges = [
    ("init_arguer", "critic_1"),
    ("init_arguer", "critic_2"),
    ("init_arguer", "critic_3"),
    ("critic_1", "final_arguer"),
    ("critic_2", "final_arguer"),
    ("critic_3", "final_arguer")
]

graph = Graph(agents, edges)
graph.process()
graph_response = graph.final_response
    
\end{lstlisting}

\end{document}

%% file: tables/case_studies.tex
\begingroup
\captionsetup{font=bf}
\setlength{\tabcolsep}{4pt}
\renewcommand{\arraystretch}{1.4}
\begin{table*}[htbp]
\caption{A summary of empirical case studies. Mechanistic fidelity studies support claims we make about how the system is operating. Efficacy checks compare the output of the system against zero-shot. One case study explores Plurals as a system for managing LLM abstentions. The leftmost column lists the study number and where to find more details. See Supplemental Materials (SM) 2 for multilevel logistic regressions of efficacy experiments.}
\label{tab:empirical-studies}
\begin{tabular}{|p{2cm}|p{2cm}|p{2.7cm}|p{7.5cm}|}
\hline
\textbf{Study No.} & \textbf{Type} & \textbf{System\newline Component(s)} & \textbf{Result} \\
\hline
1 (Appendix \ref{appendix_diversity}) & Mechanistic\newline fidelity & Personas & Using ANES personas yields more diverse responses over single-attribute personas (100\% of comparisons for Claude Sonnet, 95\% of comparisons for GPT-4o). \\
\hline
2 (Appendix \ref{appendix_annotation}) & Mechanistic\newline fidelity & Combination\newline instructions & We developed instructions based on democratic deliberation literature. The fidelity of (a subset of) these instructions was validated by crowdworkers (89\% accuracy when comparing the model's output to the given instructions). \\
\hline
3 (SM 3) & Efficacy & Personas, Ensembles, Moderators & Conservatives preferred solar panel company ideas from a simulated focus group of conservatives over zero-shot generation in 88\% of trials.\\
\hline
4 (SM 4) & Efficacy & Personas, DAGs & Liberals preferred charter school ideas from a simulated focus group of liberals over chain-of-thought zero-shot generation in 69\% of trials. \\
\hline
5 (SM 5) & Efficacy & Personas, DAGs & Liberals preferred homeless shelter proposals from a simulated focus group of liberals over chain-of-thought zero-shot generation in 66\% of trials. \\
\hline
6 (Appendix \ref{mod_study}) & Moderation & Moderators & Using Plurals, end-users can create steerable LLM guardrails (91\% accuracy in a value-based abstention experiment). \\
\hline
\end{tabular}
\end{table*}
\endgroup

%% file: tables/moderation_table.tex
\begin{table}[h]
\caption{Classifications for moderation experiment. Moderators were initialized with different harm concerns, and told to reject tasks \textit{if and only if} these tasks violated the specific harm they were to defend against.}
\label{tab:moderation_table}
\begin{tabular}{llrr}
\toprule
 & classification & accept & reject \\
\textbf{value} & \textbf{harm} &  &  \\
\midrule
\multirow[t]{2}{*}{environmental} & environmental & 0 & 90 \\
 & physical & 90 & 0 \\
\cline{1-4}
\multirow[t]{2}{*}{physical} & environmental & 86 & 4 \\
 & physical & 28 & 62 \\
\cline{1-4}
\bottomrule
\end{tabular}
\end{table}

%% file: tables/ideals.tex
\onecolumn
\section{Deliberation Ideals}
\begingroup
\captionsetup{font=bf}
\setlength{\tabcolsep}{6pt}
\renewcommand{\arraystretch}{1.4}

\begin{longtable}{|p{2.5cm}|p{2.5cm}|p{2.5cm}|p{4cm}|p{4cm}|}
\caption{Translating ideals of deliberative democracy into instructions for LLMs. Starting from the taxonomy in Bächtiger et al. \cite{bachtiger_deliberative_2018}, two authors engaged in an iterative process where we first screened ideals for relevance to AI agents and then translated ideals into LLM instructions.}
\label{tab:deliberation_ideals} \\
\hline
{\raggedright \textbf{First Generation Ideal}\par}
& {\raggedright \textbf{Second Generation Ideal}\par}
& {\raggedright \textbf{Inclusion}\par}
& {\raggedright \textbf{First Generation Instructions}\par}
& {\raggedright \textbf{Second Generation Instructions}\par} \\
\hline
\endfirsthead

\multicolumn{5}{c}{\tablename\ \thetable{} -- Continued from previous page} \\
\hline
{\raggedright \textbf{First Generation Ideal}\par}
& {\raggedright \textbf{Second Generation Ideal}\par}
& {\raggedright \textbf{Inclusion}\par}
& {\raggedright \textbf{First Generation Instructions}\par}
& {\raggedright \textbf{Second Generation Instructions}\par} \\
\hline
\endhead

\hline
\multicolumn{5}{r}{\textit{Continued on next page}} \\
\endfoot

\hline
\endlastfoot

{\raggedright Respect\par}
& {\raggedright Unrevised\par}
& {\raggedright \textbf{YES.}\par}
& {\raggedright Respect each other's viewpoints.\par}
& {\raggedright Respect each other's viewpoints.\par} \\
\hline

{\raggedright Absence of power\par}
& {\raggedright Unrevised\par}
& {\raggedright \textbf{NO.} In the current implementation, Agents do not necessarily see the identities of other Agents, so this attribute is N/A.\par}
& {\raggedright —\par}
& {\raggedright —\par} \\
\hline

{\raggedright Equality\par}
& {\raggedright Inclusion, mutual respect, equal communicative freedom, equal opportunity for influence\par}
& {\raggedright \textbf{NO.} We design Structures specifically to upweight certain voices, nullifying equality.\par}
& {\raggedright —\par}
& {\raggedright —\par} \\
\hline

{\raggedright Reasons\par}
& {\raggedright Relevant considerations\par}
& {\raggedright \textbf{YES.}\par}
& {\raggedright Give more weight to rational arguments rather than emotional ones.\par}
& {\raggedright Use empathy when engaging with others. Give value to emotional forms of communication, such as narrative, rhetoric, testimony, and storytelling.\par} \\
\hline

{\raggedright Aim and consensus\par}
& {\raggedright Aim at both consensus and clarifying conflict\par}
& {\raggedright \textbf{YES.}\par}
& {\raggedright Use rational-critical debate to arrive at a consensus.\par}
& {\raggedright Work to understand where every party is coming from. The goal is clarifying conflict, not necessarily resolving it.\par} \\
\hline

{\raggedright Common good orientation\par}
& {\raggedright Orientation to both common good and self-interest constrained by fairness\par}
& {\raggedright \textbf{YES.}\par}
& {\raggedright Aim to achieve the common good.\par}
& {\raggedright Aim to achieve the common good. It is okay to aim for self-interest if this is constrained by fairness.\par} \\
\hline

{\raggedright Publicity\par}
& {\raggedright Publicity in many conditions, but not all (e.g. in negotiations when representatives can be trusted)\par}
& {\raggedright \textbf{NO.} The notion of publicity is not applicable to AI agents.\par}
& {\raggedright —\par}
& {\raggedright —\par} \\
\hline

{\raggedright Accountability\par}
& {\raggedright Accountability to constituents when elected, to other participants and citizens when not elected\par}
& {\raggedright \textbf{NO.} Because Agents do not make decisions, they cannot be accountable.\par}
& {\raggedright —\par}
& {\raggedright —\par} \\
\hline

{\raggedright Sincerity\par}
& {\raggedright Sincerity in matters of importance; allowable insincerity in greetings, compliments, and other communications intended to increase sociality\par}
& {\raggedright \textbf{NO.} AI agents do not have notions of sincerity.\par}
& {\raggedright —\par}
& {\raggedright —\par} \\
\end{longtable}
\endgroup

%% file: tables/model_summary.tex
\begin{table}[!htbp] \centering 
  \caption{Mixed effect logistic results from efficacy studies. Participants chose between Plurals or non-Plurals output. The outcome variable is choosing Plurals. Models 1-4 have a random intercept for participants. Model 4 collapses across studies. The fixed effect intercept represents the odds (exponentiated logit coefficient) of choosing our system for a typical participant.} 
  \label{summary} 
\begin{tabular}{@{\extracolsep{5pt}}lcccc} 
\\[-1.8ex]\hline 
\hline \\[-1.8ex] 
 & \multicolumn{4}{c}{Dependent Variable: Plurals Option Chosen} \\ 
\cline{2-5} 
 & Solar & School & Housing & Overall \\ 
\\[-1.8ex] & (1) & (2) & (3) & (4)\\ 
\hline \\[-1.8ex] 
 Constant & 15.631 & 3.932 & 2.812 & 5.855 \\ 
  & t = 5.559$^{***}$ & t = 2.466$^{**}$ & t = 2.518$^{**}$ & t = 5.734$^{***}$ \\ 
  & & & & \\ 
\hline \\[-1.8ex] 
Random Intercept Variance (Person) & 2.501 & 5.178 & 2.503 & 4.043 \\ 
Observations & 300 & 300 & 200 & 800 \\ 
Log Likelihood & $-$93.969 & $-$139.743 & $-$109.423 & $-$347.845 \\ 
Akaike Inf. Crit. & 191.937 & 283.486 & 222.846 & 699.690 \\ 
Bayesian Inf. Crit. & 199.345 & 290.894 & 229.443 & 709.059 \\ 
\hline 
\hline \\[-1.8ex] 
\textit{Note:}  & \multicolumn{4}{r}{$^{*}$p$<$0.1; $^{**}$p$<$0.05; $^{***}$p$<$0.01} \\ 
\end{tabular} 
\end{table} 